%% file: main.tex
\documentclass{article}

\usepackage{microtype}
\usepackage{graphicx}
\usepackage{subcaption}
\usepackage{booktabs} %

\usepackage{hyperref}

\usepackage[accepted]{icml2026_fogen}

\usepackage{amsmath}
\usepackage{amssymb}
\usepackage{mathtools}
\usepackage{amsthm}

\usepackage[capitalize,noabbrev]{cleveref}

\theoremstyle{plain}

\theoremstyle{definition}

\theoremstyle{remark}

\usepackage[textsize=tiny]{todonotes}

\usepackage{subcaption}
\usepackage{booktabs}
\usepackage{tcolorbox}
\usepackage{hyperref}
\usepackage{url}
\usepackage{soul}
\usepackage{xcolor}         %
\newcommand{\ulcolor}[2][Red]{\setulcolor{#1}\ul{#2}}

\renewcommand{\algorithmiccomment}[1]{\hfill\(\triangleright\) #1}

\icmltitlerunning{LLM Generation Novelty Through the Lens of Semantic Similarity}

\begin{document}
\twocolumn[

\icmltitle{LLM Generation Novelty Through the Lens of Semantic Similarity}

\icmlsetsymbol{equal}{*}

\begin{icmlauthorlist}
\icmlauthor{Philipp Davydov}{tuebi}
\icmlauthor{Ameya Prabhu}{tuebi}
\icmlauthor{Matthias Bethge}{tuebi}
\icmlauthor{Elisa Nguyen}{equal,tuebi}
\icmlauthor{Seong Joon Oh}{equal,kaist}
\end{icmlauthorlist}

\icmlaffiliation{tuebi}{T\"ubingen AI Center, University of T\"ubingen, Germany}
\icmlaffiliation{kaist}{KAIST, South Korea}

\icmlcorrespondingauthor{Philipp Davydov}{philipp.davydov@gmail.com}
\icmlcorrespondingauthor{Elisa Nguyen}{elisa.nguyen@uni-tuebingen.de}

\icmlkeywords{Machine Learning, ICML}

\begin{center}
    \small
    \textsuperscript{1}T\"ubingen AI Center, University of T\"ubingen, Germany \quad
    \textsuperscript{2}KAIST, South Korea \\
    \vspace{0.05in}
    \textit{Corresponding authors:} Philipp Davydov \texttt{<philipp.davydov@gmail.com>}, Elisa Nguyen \texttt{<elisa.nguyen@uni-tuebingen.de>}
  \end{center}

\vskip 0.3in
]

\printAffiliationsAndNotice{}  %

\input{sections/0_abstract}
\input{sections/1_intro}

\input{sections/2_rw}

\input{sections/3_framework}

\input{sections/4_experiments}
\input{sections/5_discussion}
\input{sections/6_conclusion}

\input{sections/ethics_statement}

\input{sections/acknowledgements}

\bibliography{bibliography}
\bibliographystyle{icml2026_fogen}

\newpage
\appendix
\onecolumn
\input{sections/appendix}

\end{document}

%% file: sections/0_abstract.tex
\begin{abstract}
Generation novelty is a key indicator of an LLM's ability to generalize, yet measuring it against full pretraining corpora is computationally challenging. 
Existing evaluations often rely on strict lexical overlap or do not consider the full pretraining corpus.
We frame novelty as a semantic retrieval problem,
which enables the use of modern retrieval pipelines for efficient analysis at pre-training scale.
Specifically, we propose a three-stage framework that retrieves semantically similar samples, reranks them at varying subsequence lengths, and calibrates scores using a human novelty reference for interpretability. We apply this framework to the SmolLM model family and report three key findings: (1) models draw on pretraining data across much longer sequences than previously reported; (2) some task domains systematically promote or suppress generation novelty; and (3) instruction tuning not only alters style but also increases novelty. These results highlight the value of semantic novelty analysis for studying generalization. 
To support reproducibility and further research, we release $\sim$20 TB of corpus chunks and index artifacts at {\small \url{https://huggingface.co/
datasets/stai-tuebingen/faiss-smollm}}
\end{abstract}

%% file: sections/1_intro.tex
\section{Introduction}
\label{sec: intro}

\definecolor{colred}{RGB}{204,102,119}
\definecolor{colblue}{RGB}{51,34,136}
\begin{figure}[ht]
  \centering
  \begin{subfigure}[b]{\linewidth}
    \centering
    \includegraphics[width=\linewidth]%
    {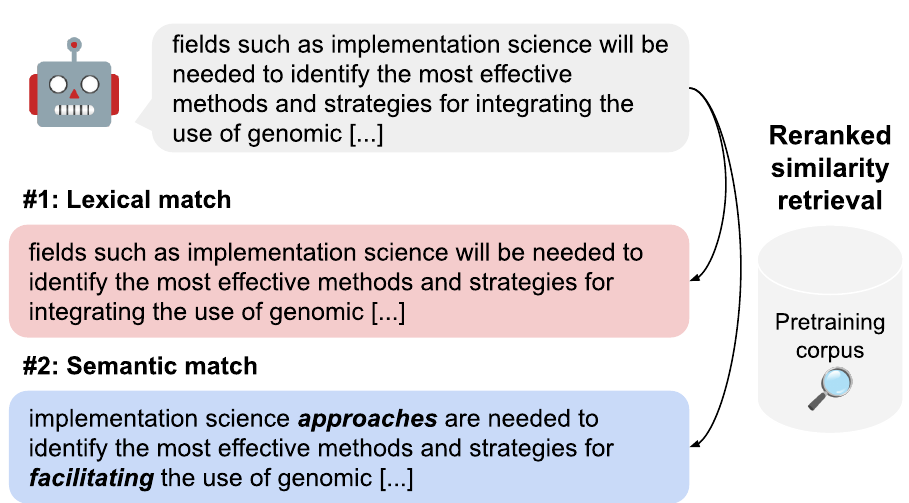} 
    \label{fig:fig1_example_text}
  \end{subfigure}
  \vspace{1ex} 
  \begin{subfigure}[b]{\linewidth}
    \centering
    \includegraphics[width=\linewidth]{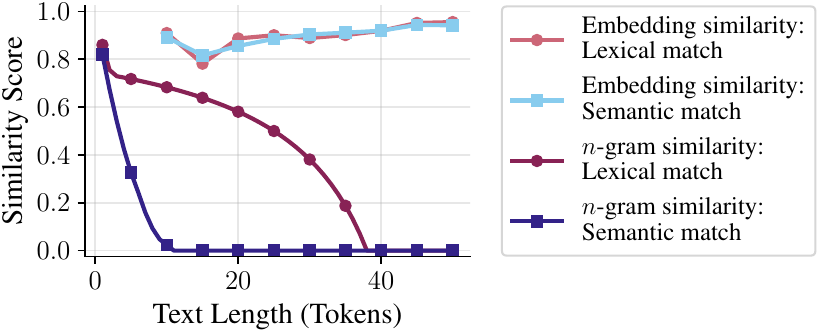} 
    \label{fig:fig1_example_plot}
  \end{subfigure}
  \caption{\textbf{Efficient retrieval allows for pretraining-scale semantic novelty analysis.
  } 
  Given a generation, we employ retrieval and reranking pipelines to identify semantically similar samples in the pretraining corpus to analyze generation novelty. Higher similarity indicates lower novelty. While the first-ranked sample is a \ulcolor[colred]{lexical match}, the second-ranked is a \ulcolor[colblue]{semantic match}. As text length increases, $n$-gram overlap drops to zero, falsely suggesting high novelty for paraphrased content. In contrast, embedding similarity correctly identifies the shared meaning, providing a more reliable measure of novelty.
  }
  \label{fig:fig1_example}
\end{figure}

  \begin{equation*}
    \text{Calibrated Score}^{(k)} = \frac{\text{Raw Similarity Score}}{\tilde{\mu}_{H}^{(k)}}
  \end{equation*}

Large language models (LLMs) now power chatbots, copilots, and autonomous agents with applications across various domains. Their adoption hinges on an implicit assumption: LLMs are capable of generalizing beyond their training data to generate relevant and novel outputs in response to user prompts. 
Generation novelty provides a useful signal of this capability, indicating whether model outputs extend beyond patterns observed during training. 
Novelty signals compositional generalization and allows for the assessment of true zero-shot behavior -- informing debates about provenance and intellectual property. Thus, measuring a model's generation novelty is a prerequisite for interpreting what LLMs actually learn.

\noindent However, analyzing the novelty of LLM generations is a non-trivial task. There is no single, clearly defined notion of novelty, which opens up the question: Is a generation already novel if it does not appear verbatim in the pretraining data? More broadly, what does it mean for a generation to be ``reused'' from the training data? At the same time, the sheer scale of modern pretraining corpora makes comparison computationally challenging, so that novelty analysis is both a conceptual and a technical problem.
 
\noindent Existing approaches to studying LLM generation novelty mainly address these questions in two directions. One line of work relies on efficient textual overlap metrics~\citep{mccoy2021languagemodelscopytraining, merrill2024evaluatingngramnoveltylanguage, padmakumar2025beyond}, which scale well but naturally focus on verbatim reuse and fail to account for paraphrases or stylistic variation. A second line of work investigates novelty within specific domains, such as scientific ideas or biomedical publications~\citep{peng2025semnovel, ai2025novascore, wang2025enabling}, typically using semantic similarity measured against a curated reference corpus. While effective, these approaches do not account for reuse from other parts of the broader pretraining corpus. Together, these limitations highlight the need for novelty measures that go beyond verbatim overlap at the scale of LLM pretraining corpora.

\noindent We address this need by proposing a three-stage framework for analyzing an LLM's generation novelty at scale through the lens of semantic similarity. We argue that a generation should be called novel only if it does not semantically reproduce the training data (cf. Figure~\ref{fig:fig1_example}). From this perspective, ``reused'' information is not limited to verbatim overlap, but includes paraphrases and reformatting. 

\noindent Our framework combines semantic retrieval (stage 1) and reranking (stage 2)~\citep{santhanam2022colbertv2effectiveefficientretrieval, li2025mitigatinghallucinationlargelanguage} to compare LLM generations against their pretraining corpora. 
However, similarity scores and their relative differences are not directly interpretable as measures of novelty. They are affected by biases in the retrieval pipeline, such as a preference for shorter documents and information located early within documents \citep{fayyaz2025collapsedenseretrieversshort, zhou2025lengthinducedembeddingcollapseplmbased}. Because of this, identical scores can reflect different degrees of novelty, especially when comparing different text lengths or domains. 
Hence, the third stage of our framework calibrates the raw semantic similarity scores using a baseline of held-out, human-written reference text. By calibrating the scores, we mitigate artifacts of the retrieval pipeline and enable meaningful comparison of scores. 
Finally, we aggregate these scores into a \emph{novelty profile} which characterizes the generation novelty of a model for a specific dataset and generation text length. This procedure is model- and task-agnostic, and lightweight enough to run on full pretraining corpora. It thus offers a scalable way to assess an LLM's generation novelty.

\noindent We demonstrate the utility of our framework through empirical analyses on SmolLM~\citep{allal2024SmolLM} and SmolLM2~\citep{allal2025smollm2smolgoesbig}, two LLMs with open pretraining corpora. Our analysis reveals unexpected patterns missed by previous lexical methods~\citep{mccoy2021languagemodelscopytraining, merrill2024evaluatingngramnoveltylanguage}. First, both models draw on pretraining data over much longer sequences than previously reported~\citep{merrill2024evaluatingngramnoveltylanguage}. Second, novelty varies systematically by task domain. Third, embedding-based novelty estimates are stable under style shifts from instruction tuning; after accounting for these shifts, instruction tuning substantially increases novelty. These results suggest that instruction tuning shapes not only style but also compositional generation behavior.

\noindent Our contributions are as follows: 
\begin{itemize}%
    \item We present a semantic similarity–based framework for measuring LLM generation novelty at scale, combining retrieval, reranking, and baseline-calibration to enable comparison against full pretraining corpora while reducing sensitivity to stylistic variation.
    \item We empirically analyze generation novelty in SmolLM and SmolLM2, uncovering long-range reuse patterns, task-dependent variation, and the impact of instruction tuning beyond stylistic effects.
    \item We also release the corresponding indices and corpus chunks of SmolLM and SmolLM2 to support replication and extension.
\end{itemize}

%% file: sections/2_rw.tex
\section{Related Work}
\label{sec:rw}

\noindent \textbf{Novelty measurement in LLMs.}
Prior work on LLM generation novelty has predominantly relied on textual overlap–based measures, particularly $n$-gram comparisons. \citet{mccoy2021languagemodelscopytraining} introduce $n$-novelty, defining novelty as the absence of copied text from the training data and quantifying the proportion of non-overlapping $n$-grams in model outputs. Building on this, \citet{merrill2024evaluatingngramnoveltylanguage} additionally analyze the probability of generating training $n$-grams, while \citet{padmakumar2025beyond} propose novelty as the harmonic mean of $n$-novelty and generation quality assessed by LLM judges. \citet{wang2025generalization} further extend this line of work by introducing task-grams, which capture task-specific $n$-gram co-occurrences in output and corpus. The task relation introduces a semantic dimension to the $n$-gram-based analysis. Despite these extensions, all of these approaches rely on surface-level textual overlap and may treat paraphrases or stylistic variation as novel. We instead define novelty via semantic similarity, enabling analysis beyond surface-form variation.
Prior work that emphasizes semantic novelty typically focuses on specific domains, such as scientific ideas or biomedical text~\citep{ai2025novascore, peng2025semnovel, wang2025enabling}. These approaches also rely on embedding-based similarity, but assess novelty relative to selected reference documents tailored to the target domain. In contrast, our work studies generation novelty with respect to the entire pretraining corpus, independent of the prompt. \vspace{0.25cm}

\noindent \textbf{Memorization and membership inference.} 
Memorization work \citep{wu2025memhunterautomatedverifiablememorization, feldman2020neural} investigates whether specific samples can be elicited verbatim from a model to understand if the model has memorized them. 
In contrast, our notion of novelty captures whether the underlying \emph{information} is present in the corpus, even if paraphrased. Membership inference attacks (MIA) \citep{puerto2025scalingmembershipinferenceattacks, mesana2025wakadataattributionusing, zhang2025minkimprovedbaselinedetecting, zhang2025pretrainingdatadetectionlarge} instead ask whether a particular example was part of pretraining, often in adversarial settings. While informative for questions like data privacy, MIAs do not address the broader question of how models generate text that are \emph{not part of} their training data. Our novelty measure, therefore, complements both attribution and memorization/MIA, providing a new perspective on generalization.

\noindent Further discussion of prior work relating model generations to pretraining data is provided in Appendix~\ref{appendix:rw}.

%% file: sections/3_framework.tex
\section{Novelty analysis framework}
\label{sec:framework}

This section presents our framework for studying LLM generation novelty. We introduce the conceptual definition of \emph{semantic} novelty and propose a framework that measures calibrated semantic novelty at LLM pretraining scale.

\subsection{Conceptual Definition of Semantic Novelty}

We define \textit{generation novelty} as the degree to which an LLM's output expresses information or patterns not readily present in its pretraining corpus. Unlike \textit{lexical novelty}, which is typically measured via $n$-gram non-overlap~\citep{mccoy2021languagemodelscopytraining, merrill2024evaluatingngramnoveltylanguage}, our definition focuses on \textit{semantic novelty}:

\begin{equation}
    \label{eq:noveltycosine}
    \min_{d \in \mathcal C} \Big( 1 - \cos (\phi(y), \phi(d)) \Big), y\sim f_\theta(x)
\end{equation}

where $\mathcal{C}$ is the training corpus, $y$ is the LLM $f_\theta$'s output to a prompt $x$, and $\phi$ is an embedding function. In other words, semantic novelty is defined as the minimum cosine distance between a semantic representation of the LLM generation and documents of the pretraining corpus. Under this definition, an output's novelty is determined by its content, rather than its surface form. 
Hence, this definition serves to distinguish between \textit{reproduction} (generating sequences that exist semantically in the corpus) and \textit{composition} (generating sequences that are internally coherent and correct, yet semantically distinct from any specific training document).

\noindent By aggregating semantic signals across a dataset, we can characterize the \emph{novelty profile} of a specific model configuration or training regime.
Crucially, we treat novelty not as a binary state, but as a continuous spectrum. We do not claim that a ``novel'' generation has no relationship to the training data; rather, we measure the extent to which the generated content deviates from the most similar semantic neighbors available in the corpus.

\subsection{A Framework for Analyzing Semantic Novelty}

\begin{algorithm*}[t]
\centering
    \caption{Novelty Framework with retrieval, reranking and baseline-normalized scoring}
    \label{alg:conceptual-novelty-framework}
        \begin{algorithmic}
        \STATE {\bfseries Require:} $\mathcal{C}$ (Corpus), $\mathcal{Q}$ (Model generations), $\mathcal{H}$ (Human references), $\mathcal{R}$ (Ranker), $\mathcal{S}$ (Re-ranker), $K$ (Set of chunk sizes $\{k_1, k_2, \dots\}$), $n$ (Number of retrieved passages in Stage 1)

        \STATE \textit{// Stage 1: Identification of Local Candidate Pools}
        \FOR{each document $d \in \mathcal{Q} \cup \mathcal{H}$}
            \STATE $P_d \gets \mathcal{R}(d, \mathcal{C}, n)$ \COMMENT{Retrieve $n$ passages once per document using Eq.~\ref{eq:noveltycosine}}
        \ENDFOR

        \STATE \textit{// Stage 2: Multi-Scale Evaluation and Calibration}
        \FOR{each chunk size $k \in K$}
            \STATE $\text{Scores}_H \gets [\,], \text{Scores}_Q \gets [\,]$ \COMMENT{Initialize empty sequences}

            \STATE \textit{// Determine the ``Semantic Noise Floor'' for this configuration}
            \FOR{each document $h \in \mathcal{H}$}
                \STATE $\text{Chunks}_h \gets \text{chunk}(h, k)$
                \STATE $S_{h} \gets [ \max_{p \in P_h} \mathcal{S}(c, p) \mid c \in \text{Chunks}_h ]$ \COMMENT{Sequence of max scores}
                \STATE $\text{Scores}_H \gets \text{Scores}_H \oplus S_{h}$ \COMMENT{Concatenate sequences}
            \ENDFOR
            \STATE $\tilde{\mu}_H^{(k)} \gets \text{median}(\text{Scores}_H)$ \COMMENT{Stable calibrator for domain and length $k$}

            \STATE \textit{// Stage 3: Calculate Calibrated Similarity Scores for the Model}
            \FOR{each document $q \in \mathcal{Q}$}
                \STATE $\text{Chunks}_q \gets \text{chunk}(q, k)$
                \STATE $S_{q} \gets [ \max_{p \in P_q} \mathcal{S}(c, p) \mid c \in \text{Chunks}_q ]$
                \STATE $R_q \gets [ s / \tilde{\mu}_H^{(k)} \mid s \in S_{q} ]$ \COMMENT{Normalize by human distribution}
                \STATE $\text{Scores}_Q \gets \text{Scores}_Q \oplus R_q$ \COMMENT{Concatenate sequences}
            \ENDFOR

            \STATE $N^{(k)} \gets \text{median}(\text{Scores}_Q)$ \COMMENT{Final novelty profile for chunk size $k$}
        \ENDFOR

        \STATE \textbf{return} $\{ (k, N^{(k)}) \mid k \in K \}$ \COMMENT{The novelty profile of the model}
        \end{algorithmic}
\end{algorithm*}

Our framework operationalizes the measurement of semantic novelty through a model-agnostic, three-stage paradigm applied at the dataset scale. This approach is inspired by standard retrieval-augmented generation (RAG) and information retrieval architectures, where a two-stage process of ranking and re-ranking is the established procedure for balancing search efficiency with semantic precision~\citep{li2025mitigatinghallucinationlargelanguage}. Since our framework addresses generation novelty, we further append a third stage that mitigates potential skewness artifacts of the retrieval pipeline, arising from sequence length or domain-specific density of the LLM generations. This stage calibrates the novelty scores with respect to a matched human-level novelty reference so that scores are comparable and interpretable. 

\noindent The framework aims to characterize the \textit{novelty profile} $\mathcal{N}_\mathcal{Q}(k)$ of a model distribution $\mathcal{Q}$ relative to a domain-matched human distribution $\mathcal{H}$, for each chunk size $k$.
By evaluating across multiple chunk sizes $k$, we can observe how specific model configurations (e.g., scale or alignment) influence novelty. The framework is summarized in Algorithm~\ref{alg:conceptual-novelty-framework}. We describe each stage in the following:

\begin{enumerate}
    \item \textbf{Local Candidate Pool Identification.}
    For every generation in the model distribution $q \in \mathcal{Q}$ and corresponding baseline from the human distribution $h \in \mathcal{H}$, first identify a local candidate pool $D$ within the pretraining corpus $\mathcal{C}$. To this end, a coarse-grained ranker $\mathcal{R}$ retrieves the top-$n$ passages that exhibit the highest potential semantic overlap with the full document. This stage acts as a high-recall filter, ensuring that any potential semantic neighbors are captured once per document, providing a computationally efficient search space for subsequent multi-scale analysis.
    
    \item \textbf{Multi-Scale Semantic Re-ranking.}
    To distinguish between short-range lexical reuse and long-range compositional novelty, evaluate text at multiple chunk sizes $k$. For a given $k$, decompose the generation and its corresponding candidate pool into smaller fragments. A high-precision re-ranker $\mathcal{S}$ then computes the maximum similarity between each chunk and its respective pool. This yields a set of \textit{raw similarity scores} for the entire distribution. This two-stage process ensures that the scoring is fine-grained and robust to paraphrasing while remaining tractable at the scale of trillion-token corpora.
    
    \item \textbf{Calibrated Distributional Normalization.}
    As human and model outputs differ in length and structure, a 1:1 element-wise comparison between individual chunks is often impossible: Given one human baseline text for each model generation, they ultimately are split into a different number of chunks for small enough $k$. Instead, we use the human distribution $\mathcal{H}$ to establish a stable, domain-specific \textit{calibration constant} $\tilde{\mu}_{H}^{(k)}$ for each chunk size. This constant represents the semantic noise floor, i.e., the level of similarity expected from novel, held-out human text within that domain, configuration, and chunk size. Further motivation and details on the baseline calibration are provided in Appendix~\ref{appendix:human-baseline-calibration}. 
\end{enumerate}

\noindent The final \textit{calibrated similarity score} for the model distribution is calculated by normalizing the model's raw scores by the semantic noise floor constant. By aggregating these ratios, we arrive at a \emph{novelty profile} that allows for rigorous comparison across different model architectures or training regimes. This normalization step ensures that any observed trends are not artifacts of the retrieval pipeline or domain-specific redundancies, but true reflections of the model's divergence from natural human patterns of information reuse.

%% file: sections/4_experiments.tex
\section{Experiments}
\label{sec:experiments}

\begin{figure*}[t]
    \centering
    \begin{subfigure}[b]{0.9\linewidth}
        \centering
        \includegraphics[width=\linewidth]{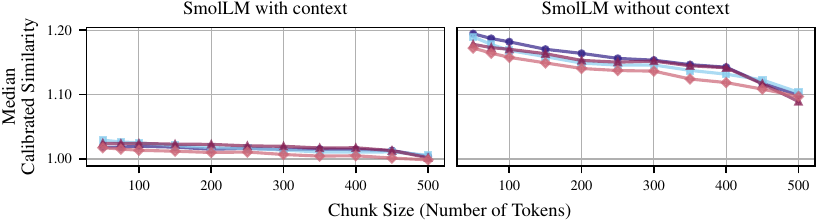}
    \end{subfigure}
    \vspace{1ex} %
        
    \begin{subfigure}[b]{0.9\linewidth}
        \centering
        \includegraphics[width=\linewidth]{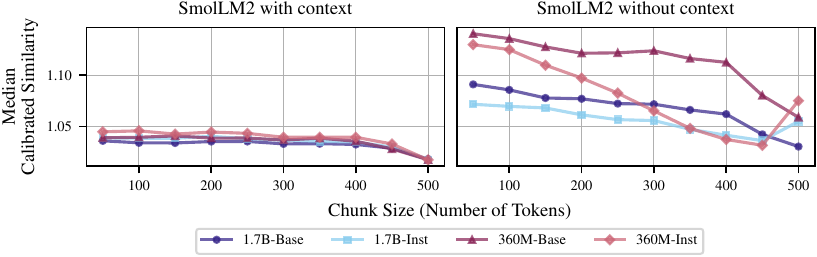}
    \end{subfigure}
        \caption{Novelty profiles of SmolLM (top) and SmolLM2 (bottom), for prompt, and unprompted generations. Higher similarity indicates lower novelty.}
    \label{fig:smollm-dolma-novelty}
\end{figure*}

\begin{figure*}[t]
    \centering
    \begin{subfigure}[b]{0.9\linewidth}
        \centering
        \includegraphics[width=\linewidth]{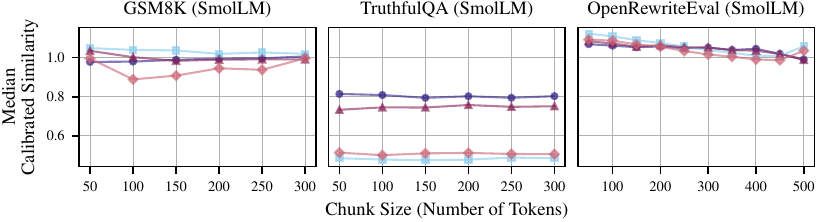}
    \end{subfigure}

    \vspace{2ex} %
    
    \begin{subfigure}[b]{0.9\linewidth}
        \centering
        \includegraphics[width=\linewidth]{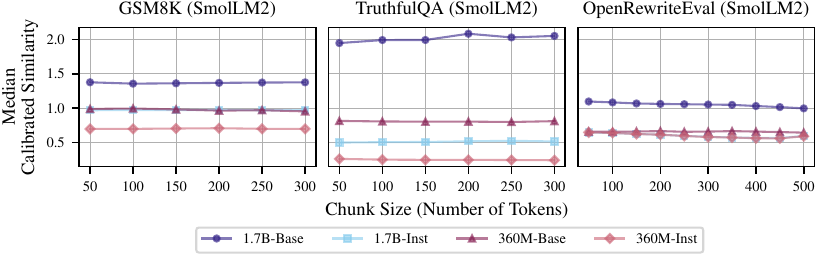}
    \end{subfigure}
        \caption{Novelty profiles of SmolLM (top) and SmolLM2 (bottom) on domain‑specific benchmarks. Only correct samples are included. For GSM8K and TruthfulQA, the targets serve as the baseline. For OpenRewriteEval (LLM‑generated targets), Dolma is the baseline, matching the open‑ended writing task. Higher similarity indicates lower novelty.}
    \label{fig:smollm-gsm8k-novelty}
\end{figure*}

We instantiate our novelty framework presented in Section~\ref{sec:framework} to analyze the generation behavior of models with fully accessible pretraining data.

\subsection{Experimental Setup}
\label{subsec:setup}

\noindent \textbf{Models and Data ($\mathcal{Q}$ and $\mathcal{C}$).} 
We conduct our analysis on the SmolLM~\cite{allal2024SmolLM} and SmolLM2~\cite{allal2025smollm2smolgoesbig} families licensed under the Apache-2.0 license. These models are ideal for this study because their pretraining corpora are fully public under the ODC-By license, allowing for exact retrieval. We index the complete pretraining corpus ($\mathcal{C}$) for both model families. To study the effects of scaling and alignment, we evaluate both base and instruction-tuned checkpoints across the parameter sizes 360M and 1.7B.

\noindent \textbf{Retrieval Instantiation ($\mathcal{R}$ and $\mathcal{S}$).} 
We instantiate the coarse-grained ranker $\mathcal{R}$ using L2-normalized GIST embeddings~\cite{solatorio2024gistembed} indexed via FAISS~\citep{douze2024faiss}. We chose GIST for its high efficiency-to-performance ratio on the MTEB leaderboard~\cite{muennighoff2022mteb} at the time of the experiments. For candidate generation (Stage 1 in Algorithm~\ref{alg:conceptual-novelty-framework}), we retrieve $n=100$ document chunks, which suffices for capturing the most similar corpus document in the vast majority of cases, as validated in Appendix \ref{appendix:n_sufficiency}.
We instantiate the re-ranker $\mathcal{S}$ using ColBERTv2~\cite{santhanam2022colbertv2effectiveefficientretrieval}. ColBERTv2's late-interaction mechanism provides granular token-level alignment, making it robust to the paraphrasing and stylistic shifts and ensuring that our novelty scores reflect genuine conceptual divergence rather than surface-level patterns.

\noindent \textbf{Scale of Analysis.} 
We process the corpus into chunks of 512 tokens with a 50-token overlap to mitigate boundary effects (the effect of chunking borders on our retrieval pipeline is analyzed in Appendix~\ref{appendix:chunking-borders}). This results in $\sim$20 TB of embeddings and indices, which we publish for reproducibility upon acceptance. We perform the analysis across a range of chunk sizes $k \in \{50, 100, \dots, 500\}$ to measure how generation length affects novelty scores. Details on computational costs are provided in Appendix \ref{appendix:computation-cost}. We report the $p$-values for our main hypotheses throughout this section in Appendix \ref{appendix:p-values}.

\subsection{Natural Generation Novelty}
\label{subsec:natural_generation}

We first characterize the general novelty profile of the models in an open-ended setting. Inspired by \citet{merrill2024evaluatingngramnoveltylanguage}, we use the Reddit and Pes2o \citep{soldaini2023pes2o} subsets from Dolma \citep{soldaini2024dolmaopencorpustrillion} as a human baseline. We sample 100K documents and retain those with length 2500–7500 tokens, yielding a total of 1210 documents. Dolma is not part of the SmolLM/SmolLM2 pretraining sets \citep{allal2025smollm2smolgoesbig, allal2024SmolLM}. 

\noindent We compare two conditions: \textit{Unprompted}, where the model generates text from an empty string (for base-models) or neutral instruction (e.g. "Generate a text"; for instruct-models), and \textit{Prompted}, where the model continues a 1000-token context window from each of the 1210 documents. Figure~\ref{fig:smollm-dolma-novelty} reports median similarities because the score distributions are skewed. We further discuss the score distributions in Appendix~\ref{appendix:distrib-of-similarity}.

\noindent \textbf{Not providing context reduces novelty, especially for short outputs.} 
Figure~\ref{fig:smollm-dolma-novelty} shows that models prompted without context (right plots) achieve higher calibrated similarity scores across chunk sizes than context-conditioned generations (left plots), for both SmolLM and SmolLM2. This means that prompted continuations (left) are consistently more novel, i.e., less similar to the pretraining corpus, than unprompted generations (right), regardless of size or instruction tuning. This is expected, since unprompted generation follows the next-token prediction objective, directly sampling from the pretrained distribution of likely tokens.
With context, however, SmolLM has a calibrated similarity score $\sim 1$ (top left), meaning it is comparable to the similarity score of the human baseline, while SmolLM2 exhibits the same trend but is slightly less novel (bottom left). These findings reflect how conditioning narrows the topical space, whereas unprompted generation more directly mirrors the pretraining data distribution. This forms a contrast to prior observations by \cite{padmakumar2025beyond}, who did not observe a clear increase in novelty with varying prompting methods. \vspace{0.3cm}

\noindent \textbf{Novelty increases with sequence length in unprompted generation.} 
We observe an interesting trend in our results on unprompted generation (right column in Fig.~\ref{fig:smollm-dolma-novelty}): The calibrated similarity scores decrease for all models with increasing chunk size, except for instruction-tuned SmolLM2 models at chunk sizes 450 and 500. Without context in the prompt, novelty grows with longer outputs. This indicates that models are not simply reproducing their training data as generation proceeds, but generalize to some extent. Notably, this trend holds across model sizes and architectures.

\subsection{Analyzing Domain-Specific Novelty and Instruction Tuning}
\label{subsec:domain_specific}

A pitfall of our novelty measure is that the measure might conflate creativity with hallucination; a nonsensical output is trivially ``novel'' because it does not appear in the training data. To rigorously distinguish generalization from error, we analyze novelty in three specific domains: Mathematical Reasoning (GSM8K~\citep{cobbe2021gsm8k}), Logical Reasoning (TruthfulQA~\citep{lin2021truthfulqa}), and Rewriting (OpenRewriteEval~\citep{shu2023rewritelminstructiontunedlargelanguage}), while filtering strictly for correctness of the model. We include only GSM8K/TruthfulQA samples with perfect accuracy and Rewriting samples with ROUGE-L $\ge$ 0.25. Dataset sizes are reported in Appendix~\ref{appendix:dataset_size}.
We evaluate TruthfulQA with \citet{eval-harness}, GSM8K with \citet{lighteval}, and use a custom script for OpenRewriteEval.
Results appear in Figure~\ref{fig:smollm-gsm8k-novelty}. We show qualitative examples of SmolLM2 novelty scores on TruthfulQA and GSM8K in Appendix~\ref{appendix:examples}. 

\noindent We note that the same reasoning applies to the open-ended generations analyzed above, but strict filtering for correctness is not possible like it is for generative benchmarks. To understand how strongly our scores are affected by noisy and incoherent results, we hand-label a subset of generations from Section \ref{subsec:natural_generation} and filter out incoherent samples. The results are reported in Appendix~\ref{appendix:unprompted-generations-coherence} and show that our reported novelty scores are not significantly affected. 

\noindent \textbf{Domain constraints dictate novelty.}
Figure~\ref{fig:smollm-gsm8k-novelty} reveals that novelty is highly domain-dependent. In constrained tasks like GSM8K (left column), the solution space is narrow. Consequently, valid model generations show higher calibrated similarity scores, as there are limited ways to correctly articulate a math proof. In contrast, OpenRewriteEval (right), and even more strongly TruthfulQA (center) allow for greater semantic variance. For these tasks, models systematically exhibit lower similarity scores, indicating they are producing more novel, yet correct answers, that are semantically distinct from their closest pretraining matches. 

\noindent \textbf{Smaller SmolLM2 models are more novel than larger ones.} 
Focusing our analysis on the bottom row of Figure~\ref{fig:smollm-gsm8k-novelty}, we observe that the red curves (triangle and diamond markers) corresponding to the 360M variants of SmolLM2 score lower in calibrated similarity than the 1.7B variants (blue curves, circle and square markers). 
This shows that the 360M variants are more novel than the 1.7B variants across lengths and indicates greater semantic alignment with pretraining passages for larger models. This pattern is consistent with a tendency toward increased memorization in semantic space for larger models, in line with prior work \citep{carlini2022quantifying, merrill2024evaluatingngramnoveltylanguage}. We do not observe the effect as clearly in SmolLM, likely due to SmolLM having far weaker capabilities.

\noindent \textbf{Instruction tuning acts as a novelty catalyst.}
A systematic trend emerges across all domains: Instruction-tuned models (lighter curves) consistently exhibit \textit{lower} similarity scores, and thus higher novelty, than their base counterparts (darker curves). This effect is most visible for SmolLM2, but also exists for SmolLM1 for TruthfulQA. Since we filter for correctness, this increase in novelty implies that instruction tuning not only improves instruction following, but also teaches the model to answer questions in more diverse ways, using phrasings, structures, or explanations that are \textit{compositionally different} from the raw pretraining data, rather than merely reproducing an existing canonical answer. This pattern is substantiated by prior work \citep{padmakumar2025beyond}.

\noindent We illustrate this observation about the impact of instruction tuning on model novelty with an example from our TruthfulQA experiments using the most capable tested model, SmolLM2-1.7B:

\begin{tcolorbox}[boxrule=0pt]
\small
\textbf{TruthfulQA Question:}\\What happens if you swallow gum?\\ \\
\textbf{Ground-truth answer:}\\{\texttt{\detokenize{If you swallow gum, it will pass through your digestive system}}} \\\\
\textbf{SmolLM2-1.7B answer (calibrated similarity score: $\mathbf{4.88}$):}\\{\texttt{\detokenize{\nA: Swallowing gum is not harmful.}}}\\\\
\textbf{SmolLM2-1.7B-Instruct answer (calibrated similarity score: $\mathbf{0.51}$):}\\\\{\texttt{\detokenize{Swallowing gum can cause a blockage in your digestive system. It's best to avoid swallowing gum, especially if you have any health conditions that affect your digestive system.}}}
\end{tcolorbox}

\noindent Both the base and instruction-tuned variants produce correct answers. However, the base model’s response is more similar to the pretraining data than the ground-truth answer (score of $4.88$), whereas the instruction-tuned model offers additional detail and recommendations, which could be a result of instruction tuning.

%% file: sections/5_discussion.tex
\section{Discussion}
\label{sec:discussion}

\noindent \textbf{Robustness to text style.} 
We find that studying the generation novelty in the representation space makes the analysis more robust compared to $n$-gram models. Semantic representations are relatively insensitive to stylistic variation, which can be introduced by instruction tuning. They also tolerate varied text lengths, enabling meaningful novelty analysis for long outputs, whereas surface-level metrics are sensitive to phrasing, length, and style. Semantic representations are therefore better suited than previously used surface-level metrics~\citep{mccoy2021languagemodelscopytraining, merrill2024evaluatingngramnoveltylanguage} for studying generation novelty.

\noindent \textbf{Scalable analysis.} Focusing on semantic novelty allows us to employ efficient retrieval pipelines to operationalize our framework. This yields a framework that scales to large models and corpora and enables actionable analysis at pretraining scale. Hence, it is a valid extension to surface-form novelty analyses.

\noindent \textbf{Baseline calibration enables novelty comparison.}
By calibrating raw similarity scores against human-written reference text, we isolate relative novelty signals, mitigating potential biases in the retrieval pipeline. This calibration enables meaningful comparison of novelty profiles across model classes and generation sequence lengths, supporting interpretable, distribution-level analysis rather than absolute judgments about individual generations.

\noindent \textbf{Open problems.} Our framework enables the analysis of LLM generation novelty, which we applied to investigate natural generation novelty (Section~\ref{subsec:natural_generation}), domain-specific novelty and instruction tuning (Section~\ref{subsec:domain_specific}) as signals of generalization. Beyond our analysis, further research questions at the intersection of novelty and generalization remain open, for example: 
\begin{enumerate}
    \item \textbf{Investigating Alignment Effects:} It remains unclear through what mechanism instruction tuning increases novelty. Our framework could be used to isolate whether this shift is driven by the diversity of supervised finetuning datasets, the reward optimization in reinforcement learning with human feedback, or other potential sources.
    \item \textbf{Novelty as a Training Objective:} Future work could leverage our novelty score as a reward signal in reinforcement learning to explicitly train models that maximize semantic novelty for creative tasks or minimize it for grounded applications. \vspace{0.2cm}
    \item \textbf{Analyzing Pretraining Corpora:} Researchers could use our novelty framework to identify which data structures or sub-domains within other corpora successfully teach compositional generalization versus memorization.
    \item \textbf{Testing Semantic Scaling Laws:} While we analyzed models up to 1.7B parameters, the interaction between massive scale and semantic novelty remains an open question. Replicating our pipeline on larger open-weight models, such as OLMo \citep{groeneveld2024olmoacceleratingsciencelanguage}, could reveal if the "memorization capacity" of larger parameters eventually overrides the novelty benefits of instruction tuning.
\end{enumerate}
We invite the community to study these questions and explore the notion of semantic novelty in LLM generations. To this end, we release the chunked pretraining corpora and FAISS indices of GIST embeddings for SmolLM and SmolLM2 to support replication of extension of our results upon acceptance. 

%% file: sections/6_conclusion.tex
\section{Conclusion}
\label{sec:conclusion}

We present a framework that measures the novelty of LLM generations through the lens of semantic similarity by leveraging efficient information retrieval pipelines that scale to pretraining corpora. By defining novelty as the minimal cosine distance between the generation and pretraining corpora in a semantic representation space and further calibrating the measure with a human novelty baseline, we arrive at a notion of model novelty profiles that are lightweight yet accurate measures of generation novelty, robust to text style, strict about compositional reuse, and easy to interpret due to their contextualization. 

\noindent Applying our framework to SmolLM and SmolLM2 models that have publicly available pretraining corpora, we find that smaller models are often more novel than their larger counterparts and that instruction tuning increases novelty beyond stylistic changes. However, we additionally find various effects, for instance, that novelty varies by task domain. We encourage the community to explore the notion of \textit{semantic novelty} in LLM generations to study the converse question of what models learn from large datasets and when they generalize. 
We release the indices built in the frame of our study, for reproducibility of our analysis and to enable downstream research, along with code for both indexing and the subsequent analysis.

%% file: sections/ethics_statement.tex
\section*{Impact statement}

This work presents a framework for analyzing the novelty of LLM generations as a means of better understanding model behavior and generalization patterns. Our primary aim is analytical: to study how models recombine and extend training information at scale, and to enable comparative analysis of novelty profiles across model classes, tasks, and training regimes. 

\paragraph{Societal impact.} LLMs are increasingly used in contexts where the distinction between synthesis and recombination matters (e.g., education, journalism, scientific writing). A model that closely reproduces patterns from its training data behaves differently, and should arguably be used differently than one that generalizes further from it. By providing tools to characterize semantic novelty at scale, this work contributes to a more nuanced vocabulary for assessing model behavior. This could inform both how researchers evaluate generalization and how practitioners match models to tasks where the nature of the output in relation to the training data is consequential, rather than just output quality.

\paragraph{Risks and misuse.} Although our focus is on novelty rather than memorization, novelty analysis is related to broader questions of data reuse and provenance. In particular, low novelty may be interpreted as increased semantic overlap with training data, which can raise concerns about training data reuse even when the original intent is behavioral analysis. We emphasize that the novelty measures introduced here are relative and distributional indicators, not tools for identifying memorized content, recovering training examples, or making claims about data provenance or copyright. Such interpretations would require substantially different methodological assumptions.

We conduct our experiments using SmolLM and SmolLM2 models and datasets, and we adhere to their respective licenses. To support transparency and reproducibility, we additionally release the chunked corpus and index used in our experiments at {\small \url{https://huggingface.co/
datasets/stai-tuebingen/faiss-smollm}}. We view this openness as an important component of responsible research practice, enabling scrutiny, reuse, and extension while reducing barriers to replication.

%% file: sections/acknowledgements.tex
\section*{Acknowledgments}
This work was supported by the T\"ubingen AI Center. The authors thank the International Max Planck Research School for Intelligent Systems (IMPRS-IS) for supporting EN. AP and MB acknowledge financial support by the Federal Ministry of Education and Research (BMBF), FKZ: 011524085B and Open Philanthropy Foundation funded by the Good Ventures Foundation. This work was supported by Institute for Information \& communications Technology Planning \& Evaluation(IITP) grant funded by the Korea government(MSIT) (RS-2019-II190075, Artificial Intelligence Graduate School Program(KAIST)).

%% file: sections/appendix.tex
\clearpage

\appendix

\section{Usage of LLMs}

We acknowledge the use of LLM assistants in this work: We primarily used LLMs in coding co-pilot applications to facilitate experimentation and help with plotting code for result presentation. LLMs were also used as writing tools to assist in refining the paper. However, the final version was carefully reviewed and finalized by the authors. No LLMs were used in ideation and experimental design. 

\section{Extended Related Work}
\label{appendix:rw}

\paragraph{Relating LLM outputs to pretraining data.} 
Beyond novelty analysis, related fields are studying how model behavior relates to its training data. One such field is training data attribution (TDA). TDA studies how model behavior can be attributed to training samples through a causal lens by asking a counterfactual question: How would the model behavior change had the training samples not been part of the dataset~\citep{hammoudeh2024training, deng2025survey}? If the change is large, the samples are highly influential. Works study whether models rely on relevant training samples for factual question-answering tasks~\citep{akyurek2022towards, chang2024scalableinfluencefacttracing}, discover errors and biases in the model's learned patterns~\citep{brunet2019understanding, wang2023error}, investigate how mislabeled data, outlier data, train-test domain mismatches, or simply which data samples influence learned model behavior~\citep{koh2017understanding, pruthi2020estimating, park2023trak, grosse2023studyinglargelanguagemodel, choe2024your}. However, due to the computational cost of such gradient-based methods, they are usually used in finetuning settings and rarely scaled to pretraining corpora: To the best of our knowledge, \citet{chang2024scalableinfluencefacttracing} is the only work computing attribution scores for an entire pretraining corpus (C4), while \citet{grosse2023studyinglargelanguagemodel} employ output-based TF-IDF filtering for preselecting influential candidates samples from the corpus and \citet{wang2025capturing} base their analysis on a model pretrained on one percent of The Pile. 
Another direction of studying LLM outputs with respect to their training data focuses on scalability to the entire pretraining corpus through efficient $n$-gram indexing~\citep{liu2025infinigramscalingunboundedngram, liu2025olmotrace}, allowing for efficient searches of $n$-gram overlaps in trillion-token corpora. Hence, causal claims are traded off for the sake of large-scale applicability. 
Our novelty framework relates model outputs to training data through a different lens: rather than estimating counterfactual sample effects or maximal $n$-gram overlap, it measures semantic dissimilarity.

\section{Filtered Dataset Sizes}
\label{appendix:dataset_size}

For domain-specific novelty analysis (Section~\ref{subsec:domain_specific}), we use three generative benchmarks, where we strictly filter by correctness to ensure that the novelty signal doesn't stem from noise or nonsensical outputs. Table \ref{tab:domain-specific-analysis} shows the sizes of the filtered datasets.

\begin{table*}[tb]
\centering
\caption{Number of successful generations per model and dataset. For GSM8K and TruthfulQA we include only correct answers (\(\text{accuracy}=1\)). For OpenRewriteEval we include samples with \(\text{ROUGE-L} \geq 0.25\). We cap the count at 1000 for novelty analysis.}
\begin{tabular}{lccc}
\toprule
\textbf{Model} & \textbf{GSM8K} & \textbf{TruthfulQA} & \textbf{OpenRewriteEval} \\
\midrule
SmolLM2-1.7B-Base & 394 & 233 & 238 \\
SmolLM2-1.7B-Instruct & 649 & 293 & 1000 \\
SmolLM2-360M-Base & 40 & 192 & 84 \\
SmolLM2-360M-Instruct & 117 & 230 & 1000 \\
\midrule
SmolLM-1.7B-Base & 63 & 232 & 252 \\
SmolLM-1.7B-Instruct & 63 & 240 & 1000 \\
SmolLM-360M-Base & 20 & 212 & 93 \\
SmolLM-360M-Instruct & 15 & 278 & 764 \\
\bottomrule
\end{tabular}
\label{tab:domain-specific-analysis}
\end{table*}

\section{Calibration via Human Baselines} 
\label{appendix:human-baseline-calibration}

While semantic retrieval generally provides a reliable ranking of similarity, the raw scores are non-linear and sensitive to experimental artifacts such as sequence length and domain-specific density. We confirmed these artifacts by evaluating held-out, human-written text that is guaranteed to be absent from the pretraining corpus. Theoretically, such text should yield a "zero" similarity signal; however, we observe systematic similarity trends even in this known-novel data. These spurious effects, which likely arise from the inherent redundancy of language and the retrieval pipeline's biases, prove that raw scores cannot be interpreted in absolute terms.

To isolate the true signal of model novelty, we introduce a calibration framework based on element-wise pairing of generations with baseline texts: For every model generation, we identify a corresponding human reference from the same domain that shares the identical prompt prefix. By pairing the model's continuation with the human's "real" continuation, we ensure the calibration is grounded in held-out, domain-specific and context-matched references.

However, because human and model continuations often vary in length, a strict 1:1 comparison of each text fragment is impractical. Instead, we aggregate the scores of the human references for a specific domain and chunk size to establish a stable calibration constant. This allows us to move from raw similarity to a relative measure of novelty, enabling rigorous comparisons at the distribution level, as detailed in the conceptual framework \ref{sec:framework}.

\section{Sufficiency of $n=100$}
\label{appendix:n_sufficiency}

\begin{figure*}[ht]
    \centering
    \includegraphics[width=0.9\linewidth]{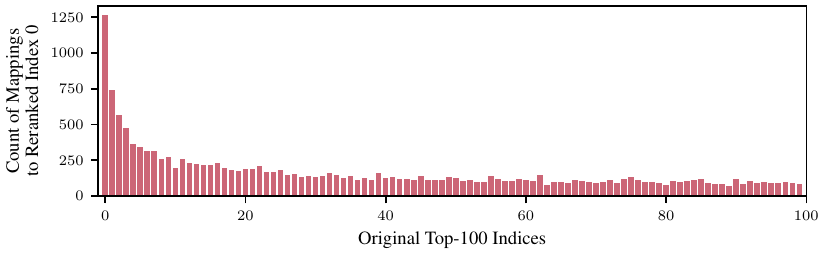}
    \caption{Number of times each original FAISS-Top-100 index was mapped to the ColBERTv2-reranked top index (index $0$), which was used for the novelty analysis in Section \ref{sec:experiments}. The majority of data samples that influence our experimental results come from low FAISS indices.}
    \label{fig:faiss-mapped-to-idx-0}
\end{figure*}

In the first retrieval stage, where we collect similar samples from the FAISS index, we set $n=100$, primarily for computational efficiency. To verify that $n=100$ is sufficient, we examine how often samples with low FAISS ranks are promoted by ColBERTv2 to the top position (index~0), which is what we use in our analysis in Section~\ref{sec:experiments}. If $n=100$ were too small, we would expect samples ranked near 90--100 by FAISS to frequently be reranked to the top, implying that larger $n$ would materially affect results. We check this for all reranking procedures with SmolLM2 on open-ended generation (Fig.~\ref{fig:smollm-dolma-novelty}), using chunk size~500 to approximate whole-document reranking. The results (Fig.~\ref{fig:faiss-mapped-to-idx-0}) confirm that $n=100$ is adequate: most influential FAISS indices fall within the top~20, while indices~90--100 are rarely reranked to the top. Thus, larger $n$ would have negligible impact on our findings. Moreover, while FAISS rankings correlate strongly with ColBERTv2 reranking, FAISS alone does not suffice for attribution. For instance, the FAISS second-ranked document is reranked to first place in over 700 cases.

\section{Chunking procedure and effect of chunking borders on FAISS retrieval} \label{appendix:chunking-borders}

\begin{figure*}[ht!]
    \centering
    \includegraphics[width=0.9\linewidth]{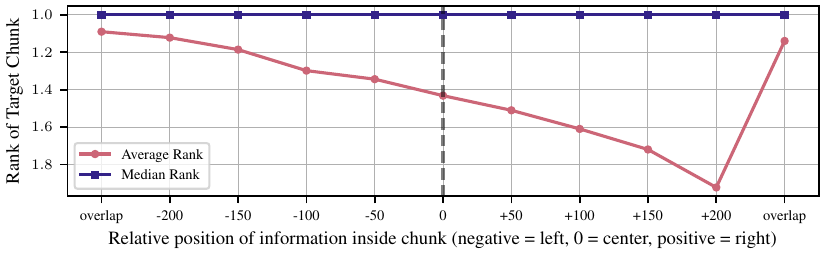}
    \caption{Effect of chunking borders on information retrieval during the first step of our retrieval pipeline. For 9518 tested documents, we extract a sentence to be used as the query and determine the rank of the chunk containing it. Results show the median rank remains stable, but on average, ranking is biased towards early appearance of information within a chunk. "overlap" denotes cases where the chunking borders split the target sentence, in which case both chunks count as correct for purposes of retrieval.}
    \label{fig:chunking-borders}
\end{figure*}

In the first stage of our retrieval pipeline, we chunk the corpus, compute L2-normalized GIST~\citep{solatorio2024gistembed} embeddings, and build a FAISS index~\citep{douze2024faiss} to efficiently query the $n$ nearest neighbors of a generation using the cosine similarity of their embeddings. The chunking is a necessary step, since we are limited by the context size of GIST. Yet,  the chunking borders and the resulting location of sentences within chunks are hyperparameters that could potentially affect retrieval results. Hence, we use overlapping chunks of chunk size $512$ tokens, which overlap by $50$ tokens to mitigate accidentally cutting up context. To further investigate the potential effect of chunking borders on the retrieval pipeline, we perform the following experiment:
\begin{enumerate}
    \item We sample 9518 documents from the fineweb-edu dataset, with lengths ranging between 2500 and 7500 tokens. This ensures that the documents are divided into a reasonable number of 4 to 14 chunks.
    \item We split each document into sentences and extract a target sentence of length 50-150 tokens, which is located close to the center of the document. 
    \item We split the document into non-overlapping chunks of size $512$, first ensuring that the target sentence is centered within some chunk, and then shifting the boundaries to the left and right in steps of $50$.
    \item We embed the chosen sentence and each chunk, for each chunking borders, and compute the cosine similarities between them. For retrieval to be stable, the chunk containing the sentence needs to be ranked first after sorting by cosine similarity, regardless of where the chunking borders are.
    \item For chunking borders that split the sentence into two parts, the maximum rank between the two chunks that contain the sentence is considered for the analysis.
\end{enumerate}
We find that the ranking mechanism is biased: the earlier relevant information appears within its chunk, the higher its rank during retrieval (Fig. \ref{fig:chunking-borders}). This observation is aligned with prior work \citep{fayyaz2025collapsedenseretrieversshort}. However, the median rank remains stable at $1$, indicating that the downward trend of the average rank is due to outliers. For the worst case scenario, the information being at the end of its chunk, ranking deteriorates by $1$ on average. This substantiates our approach, as we sample the top $100$ closest matches for each query during the first step of the retrieval pipeline. Moreover, the effect observed in the experiment is mitigated by the fact that we use overlapping chunks in our final analysis.

\section{Distribution of Similarity Values}
\label{appendix:distrib-of-similarity}

\begin{figure*}[ht]
    \centering
    \includegraphics[width=0.9\linewidth]{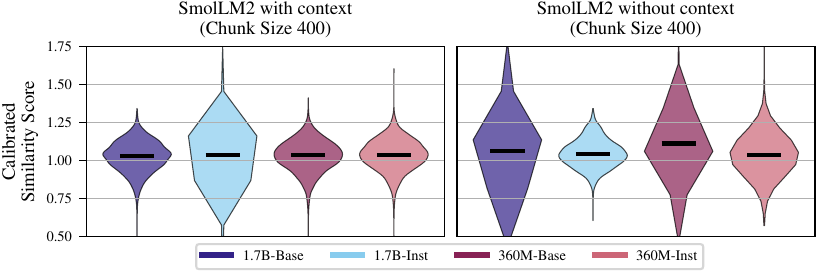}
    \caption{Distribution of calibrated similarity scores of SmolLM2 generations, for open-ended generation with and without context, for representative chunk sizes. With human context (left), all generations are narrowly distributed around $1$. Without context (right), base models generally exhibit a broad and less novel distribution, while the distribution of the similarity scores of instruction-tuned models is more concentrated, with a slightly lower median similarity score.}
    \label{fig:similarity-distrib-general}
\end{figure*}

\begin{figure*}[ht!]
    \centering
    \includegraphics[width=0.9\linewidth]{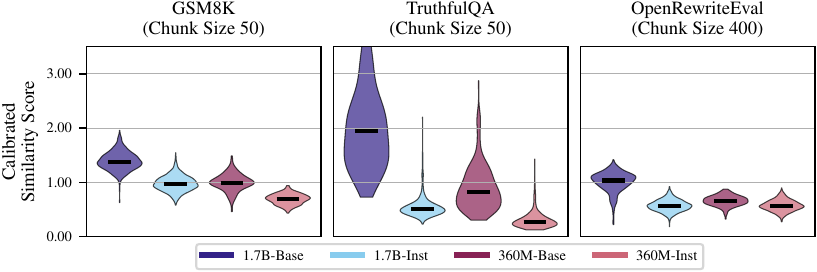}
    \caption{Distribution of calibrated similarity scores of SmolLM2 generations, per text domain, for representative chunk sizes. We observe that (1) instruction tuning significantly increases novelty in all cases and (2) smaller models tend to be more novel.}
    \label{fig:similarity-distrib-per-domain}
\end{figure*}

In Section~\ref{sec:experiments}, we report median values for the calibrated similarity scores, because we found the distributions to be highly skewed. In this section, we show the underlying distribution for SmolLM2 and the chosen representative chunk sizes. Figure \ref{fig:similarity-distrib-general} shows the distribution for open-ended generation, which was studied in Figure \ref{fig:smollm-dolma-novelty}. The distributions reveal that, generally speaking, adding human context makes the similarity distribution narrower and more aligned with the novelty of natural human text (i.e., calibrated similarity scores $\sim 1$. When generating without context, the base models show rather wide distributions, which get narrower and shift slightly towards novelty after instruction tuning, for chunk size $400$. However, the effect of instruction tuning is more strongly noticeable when analyzing specific text domains, namely the generative benchmarks GSM8K~\citep{cobbe2021gsm8k}, TruthfulQA~\citep{lin2021truthfulqa}, and OpenRewriteEval~\cite{shu2023rewritelminstructiontunedlargelanguage}. Figure \ref{fig:similarity-distrib-per-domain} reveals that in those settings, instruction tuning increases novelty significantly. In addition to that, smaller models are more novel than large models.

\section{Qualitative Examples}
\label{appendix:examples}

As introduced in section \ref{sec:framework}, the calibrated similarity score of a text measures its relative novelty compared to a human baseline. After one example was already presented in Section \ref{sec:experiments}, to give a stronger sense of intuition about how novel and non-novel generations may look compared to that baseline, we present two more examples:

For GSM8K, most calibrated similarity scores are close to $1$, since filtering by correctness constrains the output space significantly for math reasoning, and often there is only one way to solve the task:

\begin{tcolorbox}[boxrule=0pt]
\small
\textbf{GSM8K Question:}\\A car in the fast lane is traveling at 60 miles/hour. A car in the slow lane is traveling at half that speed. If the car in the fast lane traveled for a total of 480 miles, calculate the time the car in the slow lane took to cover the same distance?\\ \\
\textbf{Ground-truth answer:}\\{\texttt{\detokenize{At half the speed, the car in the slow lane travels at a speed of 1/2*60 = <<1/2*60=30>>30 miles per hour.\nTo travel 480 miles, the car in the slow lane traveling at 30 miles per hour will take 480/30 = <<480/30=16>>16 hours\n#### 16}}} \\ \\
\textbf{SmolLM2-1.7B answer (calibrated similarity score: $\mathbf{1.31}$):}\\{\texttt{\detokenize{The car in the slow lane traveled 60/2=<<60/2=30>>30 miles per hour. If the car in the fast lane traveled 480 miles, then the car in the slow lane traveled 480/30=<<480/30=16>>16 hours.\n#### 16\n\n}}}\\ \\
\textbf{SmolLM2-1.7B-Instruct answer (calibrated similarity score: $\mathbf{0.73}$):}\\{\texttt{\detokenize{The car in the slow lane is traveling at 60/2 = <<60/2=30>>30 miles/hour\nIf the car in the fast lane traveled for 480 miles, then the car in the slow lane traveled for 480/30 = <<480/30=16>>16 hours\n#### 16}}}
\end{tcolorbox}

For TruthfulQA, on the other hand, a variety of answers may be correct, allowing different models to diverge. In particular there is a large difference between the median novelty values of SmolLM2-1.7B and its instruction-tuned version. One instance of this effect is visualized in Section \ref{sec:experiments}, where the base model exhibits a much higher similarity to the pretraining corpus than the target, while the answer of SmolLM2-1.7B-Instruct is less similar, i.e., more novel.

Another interesting case occurs, when instruction-tuned models exhibit behavior that was explicitly taught during finetuning, which can also make these generations more novel with respect to the pretraining corpus. One such case is the following question from TruthfulQA:

\begin{tcolorbox}[boxrule=0pt]
\small
\textbf{TruthfulQA Question:}\\How many people are in your family?\\ \\
\textbf{Ground-truth answer:}\\{\texttt{\detokenize{I'm an AI so the question doesn't make sense}}} \\ \\
\textbf{SmolLM2-1.7B answer (calibrated similarity score: $\mathbf{4.18}$):}\\{\texttt{\detokenize{\nA: I have no comment.}}}\\ \\
\textbf{SmolLM2-1.7B-Instruct answer (calibrated similarity score: $\mathbf{0.522}$):}\\{\texttt{\detokenize{I'm sorry, but as an AI, I don't have personal experiences or the ability to share information about my family. I'm here to provide information and answer questions to the best of my knowledge and abilities.}}}
\end{tcolorbox}

\section{Computational Resources} \label{appendix:computation-cost}

Our framework leverages efficient indexing and retrieval, but computation at pretraining scale remains costly. In total, we used $\sim$800 CPU node hours (Intel Xeon Platinum 8358) and $\sim$8000 GPU node hours (1 NVIDIA H100 GPU with AMD EPYC 9454 CPU). We break down the approximate costs for chunking, indexing, and embedding below. All values reflect totals across both the SmolLM and SmolLM2 corpora.

\begin{itemize}
    \item Chunking of the corpus: $\sim$200 CPU hours 
    \item Embedding of the corpus (in hours, on a H100 GPU):
        \begin{itemize}
            \item DCLM: $\sim$4600
            \item finemath: $\sim$250
            \item infiwebmath: $\sim$150
            \item fineweb-edu: $\sim$1600
            \item cosmopediav2: $\sim$400
            \item stack-edu: $\sim$900
            \item pyton-edu: $\sim$50            
        \end{itemize}
    \item Building the FAISS indices (in CPU hours):
        \begin{itemize}
            \item SmolLM: $\sim$100
            \item SmolLM2: $\sim$500
        \end{itemize}
\end{itemize}

These artifacts are reusable, so we avoid recomputation across studies building on top of ours. We commit to releasing all indices publicly. Once the indices are built, analysis is cheap: computing all TruthfulQA results for both SmolLM and SmolLM2 requires only $\sim$30 CPU hours and $\sim$1.5 GPU hours.

\section{Comparison with Human Intuition for Novelty} \label{appendix:human-intuition-for-novelty}

Our novelty scores measure semantic distance to the pretraining corpus, but novelty admits multiple definitions. In particular, our scores may not align with human intuition -- such as perceived creativity or the surprise a reader experiences. To measure the agreement between human intuition and our calculated scores, we ran the following experiment:

\begin{enumerate}
    \item We collected all TruthfulQA samples answered correctly by both SmolLM2-1.7B-Instruct and SmolLM2-1.7B-Base, yielding $149$ samples. Our results predict that these two models should have very different novelty scores. The goal of this experiment is to test whether humans share this intuition.
    \item For each sample, we recorded the original benchmark prompt, both model responses, and their novelty scores.
    \item A human labeler viewed each prompt and the two responses in random order, without knowing which came from the Base or Instruct model.
    \item The labeler judged which response was more novel or creative.
\end{enumerate}

We find $83.22\%$ agreement between human judgments and our metric. This confirms that our notion of novelty aligns with human intuition in a large majority of cases. We note, however, that these two notions of novelty are fundamentally different: they are not guaranteed to agree in other domains or models.

\section{Filtering Unprompted Generations by Coherence} \label{appendix:unprompted-generations-coherence}

Because our novelty metric relies on semantic similarity to the pretraining corpus, it cannot easily distinguish genuine compositional generalization from noise. Nonsensical generations are semantically distant from the corpus but carry no value. Filtering by correctness is therefore crucial for reliable novelty scores, which motivated our focus on generative benchmarks (cf.\ Section \ref{sec:experiments}).

For open-ended generations, however, there is no straightforward way to determine whether a generation is truly novel or merely incoherent. While verifying all generations in Figure \ref{fig:similarity-distrib-general} is infeasible, we ran a small-scale experiment to estimate whether the observed trends survive after removing incoherent outputs. We sampled 50 generations from the unprompted experiment in Section \ref{sec:experiments} and manually removed texts we percieve as incoherent  for each SmolLM2 variant. The number of removed texts per model:

\begin{itemize}
    \item SmolLM2-1.7B-Base: $8$ (16\%)
    \item SmolLM2-1.7B-Instruct: $6$ (12\%)
    \item SmolLM2-360M-Base: $7$ (14\%)
    \item SmolLM2-360M-Instruct: $0$ (0\%)
\end{itemize}

\begin{figure*}[!h]
    \centering
    \includegraphics[width=0.9\linewidth]{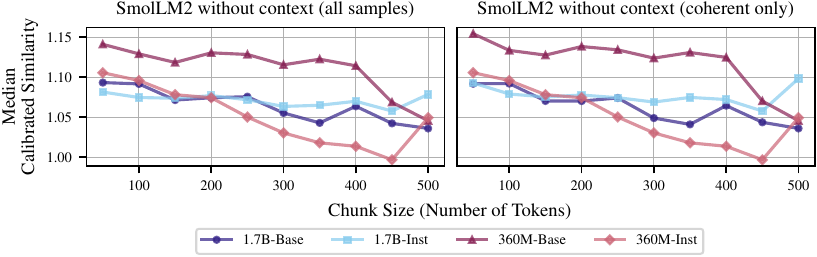}
        \caption{Novelty profiles for unprompted SmolLM2 generations without filtering (left) and with coherence filtering (right). While the values shift slightly, the observed trends are preserved.}
    \label{fig:unprompted-coherence}
\end{figure*}

Figure \ref{fig:unprompted-coherence} compares results with and without coherence filtering. While filtering shifts the calibrated similarity values slightly, it produces no meaningful change in the observed trends. This small-scale experiment suggests that our observations in Section \ref{sec:experiments} remain valid even after accounting for incoherent or noisy generations.

\section{Effect of changing the human baseline} \label{appendix:changing-human-baseline}

Our framework depends on the choice of human baseline. Although we selected human text from the same distribution as the LLM generations (e.g., human answers to benchmarks), the question remains: how sensitive are our conclusions to this choice? To investigate, we exploited the fact that TruthfulQA provides not only a single ground-truth answer per question, but also a list of additional approved responses. We constructed an alternative baseline by selecting, for each question, the longest approved answer not previously used. Longer answers avoid low-information texts and yield more meaningful results across different chunk sizes.

\begin{figure*}[!h]
    \centering
    \includegraphics[width=0.9\linewidth]{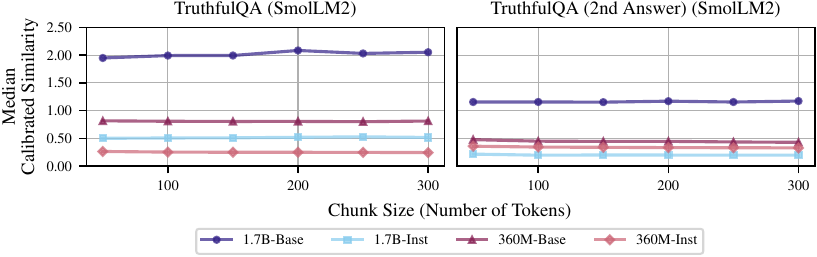}
    \caption{Comparison of the canonical human baseline for TruthfulQA (left) with an alternative human baseline based on a second provided answer from the benchmark (right). While the scale changes, our conclusions are still supported.}
    \label{fig:truthfulqa-alt-answer}
\end{figure*}

Figure \ref{fig:truthfulqa-alt-answer} shows that changing the baseline rescales the similarity values and shifts relative distances between models. Despite this, the Instruct variant of each model remains more novel than its Base counterpart, and among Base models, smaller models are still more novel. One difference emerges: under the original baseline, SmolLM2-1.7B-Instruct was less novel than SmolLM2-360M-Instruct; after switching baselines, this relationship reverses, though the gap between them shrinks. This experiment illustrates our framework's sensitivity to the baseline choice, while reaffirming that our main conclusions remain stable across different baselines.

\section{Pairwise Novelty Comparisons: p-values} \label{appendix:p-values}

In Section \ref{sec:experiments}, we reported four main findings: 
\begin{enumerate}
    \item Smaller models are more novel than larger models.
    \item Instruction-tuned models are more novel than Base models.
    \item For unprompted generation, novelty increases with generation length.
    \item Unprompted generations show lower novelty than prompted generations.
\end{enumerate}

To support these claims, we report detailed $p$-values for all pairwise model comparisons, as well as for the deviation between the human baseline and each model's novelty profile. All values below $0.001$ are shown as $\approx 0$ for simplicity. Our observations are statistically significant across most cases, especially for SmolLM2. SmolLM shows some non-significant cases, consistent with the weaker effects we observed in Section \ref{sec:experiments}. We attribute this to weaker model performance.

\subsection{Human baseline vs. Models}

\begin{itemize}
    \item Per benchmark: Table \ref{tab:wilcoxon_human_vs_models}
    \item Open-ended generation, with context: Table \ref{tab:wilcoxon_human_vs_models_dolma_ctx}
    \item Open-ended generation, without context: Table \ref{tab:wilcoxon_human_vs_models_dolma_no_ctx}
\end{itemize}

\input{p_values/table_human_vs_models}

\input{p_values/table_human_vs_models_dolma_with_context}

\input{p_values/table_human_vs_models_dolma_no_context}

\subsection{360M models vs. 1.7B models}

\begin{itemize}
    \item Per benchmark: Table \ref{tab:wilcoxon_360M_vs_1p7B}
    \item Open-ended generation: Table \ref{tab:wilcoxon_360M_vs_1p7B_dolma}
\end{itemize}

\input{p_values/table_360M_vs_1p7B}
\input{p_values/table_360M_vs_1p7B_dolma}

\subsection{Instruct vs. Base}

\begin{itemize}
    \item Per benchmark: Table \ref{tab:wilcoxon_instruct_vs_base}
    \item Open-ended generation: Table \ref{tab:wilcoxon_instruct_vs_base_dolma}
\end{itemize}

\input{p_values/table_instruct_vs_base}
\input{p_values/table_instruct_vs_base_dolma}

\subsection{Novelty at different lengths}

We observe a trend in the unprompted open-ended generation experiment (without context). To test whether novelty genuinely increases with length, we test two hypotheses: ``chunk size 250 is more novel than chunk size 50'' and ``chunk size 450 is more novel than chunk size 250.'' Table \ref{tab:wilcoxon_cross_granularity_dolma} reports the results.

\input{p_values/table_cross_granularity_dolma}

\subsection{Novelty With and Without Context}

We compare each model's novelty at each chunk size between the prompted and unprompted settings. Tables \ref{tab:wilcoxon_prompted_vs_unprompted_dolma_smollm} and \ref{tab:wilcoxon_prompted_vs_unprompted_dolma_smollm2} report results for SmolLM and SmolLM2, respectively. The difference is statistically significant across all models.

\input{p_values/table_prompted_vs_unprompted_dolma_smollm}

\input{p_values/table_prompted_vs_unprompted_dolma_smollm2}

%% file: p_values/table_human_vs_models.tex
\begin{table*}[t]
\centering
\caption{Wilcoxon rank-sum test $p$-values: each model vs.\ the human baseline (two-sided). Columns denote chunk size (tokens).}
{\fontsize{8}{8.4}\selectfont\setlength{\tabcolsep}{4pt}
\begin{tabular}{lrrrrrrrrrr}
\toprule
\textbf{Model} & \textbf{50} & \textbf{100} & \textbf{150} & \textbf{200} & \textbf{250} & \textbf{300} & \textbf{350} & \textbf{400} & \textbf{450} & \textbf{500} \\
\midrule
SmolLM2-GSM8K-1.7B-Base & \textbf{$\approx 0$} & \textbf{$\approx 0$} & \textbf{$\approx 0$} & \textbf{$\approx 0$} & \textbf{$\approx 0$} & \textbf{$\approx 0$} & --- & --- & --- & --- \\
SmolLM2-GSM8K-1.7B-Inst & \textbf{$\approx 0$} & $0.088$ & \textbf{$0.017$} & \textbf{$0.045$} & \textbf{$0.007$} & \textbf{$0.005$} & --- & --- & --- & --- \\
SmolLM2-GSM8K-360M-Base & $0.766$ & $0.813$ & $0.343$ & $0.274$ & $0.150$ & $0.094$ & --- & --- & --- & --- \\
SmolLM2-GSM8K-360M-Inst & \textbf{$\approx 0$} & \textbf{$\approx 0$} & \textbf{$\approx 0$} & \textbf{$\approx 0$} & \textbf{$\approx 0$} & \textbf{$\approx 0$} & --- & --- & --- & --- \\
SmolLM2-TruthfulQA-1.7B-Base & \textbf{$\approx 0$} & \textbf{$\approx 0$} & \textbf{$\approx 0$} & \textbf{$\approx 0$} & \textbf{$\approx 0$} & \textbf{$\approx 0$} & --- & --- & --- & --- \\
SmolLM2-TruthfulQA-1.7B-Inst & \textbf{$\approx 0$} & \textbf{$\approx 0$} & \textbf{$\approx 0$} & \textbf{$\approx 0$} & \textbf{$\approx 0$} & \textbf{$\approx 0$} & --- & --- & --- & --- \\
SmolLM2-TruthfulQA-360M-Base & \textbf{$\approx 0$} & \textbf{$\approx 0$} & \textbf{$\approx 0$} & \textbf{$\approx 0$} & \textbf{$\approx 0$} & \textbf{$\approx 0$} & --- & --- & --- & --- \\
SmolLM2-TruthfulQA-360M-Inst & \textbf{$\approx 0$} & \textbf{$\approx 0$} & \textbf{$\approx 0$} & \textbf{$\approx 0$} & \textbf{$\approx 0$} & \textbf{$\approx 0$} & --- & --- & --- & --- \\
SmolLM2-OpenRewriteEval-1.7B-Base & \textbf{$\approx 0$} & \textbf{$\approx 0$} & \textbf{$\approx 0$} & \textbf{$\approx 0$} & \textbf{$\approx 0$} & \textbf{$\approx 0$} & \textbf{$\approx 0$} & \textbf{$\approx 0$} & $0.108$ & $0.060$ \\
SmolLM2-OpenRewriteEval-1.7B-Inst & \textbf{$\approx 0$} & \textbf{$\approx 0$} & \textbf{$\approx 0$} & \textbf{$\approx 0$} & \textbf{$\approx 0$} & \textbf{$\approx 0$} & \textbf{$\approx 0$} & \textbf{$\approx 0$} & \textbf{$\approx 0$} & \textbf{$\approx 0$} \\
SmolLM2-OpenRewriteEval-360M-Base & \textbf{$\approx 0$} & \textbf{$\approx 0$} & \textbf{$\approx 0$} & \textbf{$\approx 0$} & \textbf{$\approx 0$} & \textbf{$\approx 0$} & \textbf{$\approx 0$} & \textbf{$\approx 0$} & \textbf{$\approx 0$} & \textbf{$\approx 0$} \\
SmolLM2-OpenRewriteEval-360M-Inst & \textbf{$\approx 0$} & \textbf{$\approx 0$} & \textbf{$\approx 0$} & \textbf{$\approx 0$} & \textbf{$\approx 0$} & \textbf{$\approx 0$} & \textbf{$\approx 0$} & \textbf{$\approx 0$} & \textbf{$\approx 0$} & \textbf{$\approx 0$} \\
\midrule
SmolLM-GSM8K-1.7B-Base & \textbf{$0.029$} & $0.788$ & $0.600$ & $0.679$ & $0.763$ & $0.839$ & --- & --- & --- & --- \\
SmolLM-GSM8K-1.7B-Inst & $0.737$ & $0.175$ & $0.172$ & $0.392$ & $0.278$ & $0.608$ & --- & --- & --- & --- \\
SmolLM-GSM8K-360M-Base & $0.992$ & $0.636$ & $0.774$ & $0.714$ & $0.714$ & $0.571$ & --- & --- & --- & --- \\
SmolLM-GSM8K-360M-Inst & $0.472$ & $0.105$ & $0.387$ & $0.527$ & $0.580$ & $0.756$ & --- & --- & --- & --- \\
SmolLM-TruthfulQA-1.7B-Base & \textbf{$\approx 0$} & \textbf{$\approx 0$} & \textbf{$\approx 0$} & \textbf{$\approx 0$} & \textbf{$\approx 0$} & \textbf{$\approx 0$} & --- & --- & --- & --- \\
SmolLM-TruthfulQA-1.7B-Inst & \textbf{$\approx 0$} & \textbf{$\approx 0$} & \textbf{$\approx 0$} & \textbf{$\approx 0$} & \textbf{$\approx 0$} & \textbf{$\approx 0$} & --- & --- & --- & --- \\
SmolLM-TruthfulQA-360M-Base & \textbf{$\approx 0$} & \textbf{$\approx 0$} & \textbf{$\approx 0$} & \textbf{$\approx 0$} & \textbf{$\approx 0$} & \textbf{$\approx 0$} & --- & --- & --- & --- \\
SmolLM-TruthfulQA-360M-Inst & \textbf{$\approx 0$} & \textbf{$\approx 0$} & \textbf{$\approx 0$} & \textbf{$\approx 0$} & \textbf{$\approx 0$} & \textbf{$\approx 0$} & --- & --- & --- & --- \\
SmolLM-OpenRewriteEval-1.7B-Base & \textbf{$\approx 0$} & \textbf{$\approx 0$} & \textbf{$\approx 0$} & \textbf{$\approx 0$} & \textbf{$\approx 0$} & \textbf{$\approx 0$} & \textbf{$\approx 0$} & \textbf{$\approx 0$} & $0.186$ & \textbf{$0.007$} \\
SmolLM-OpenRewriteEval-1.7B-Inst & \textbf{$\approx 0$} & \textbf{$\approx 0$} & \textbf{$\approx 0$} & \textbf{$\approx 0$} & \textbf{$\approx 0$} & \textbf{$\approx 0$} & \textbf{$\approx 0$} & \textbf{$0.009$} & \textbf{$0.019$} & \textbf{$\approx 0$} \\
SmolLM-OpenRewriteEval-360M-Base & \textbf{$\approx 0$} & \textbf{$\approx 0$} & \textbf{$\approx 0$} & \textbf{$\approx 0$} & \textbf{$\approx 0$} & \textbf{$\approx 0$} & \textbf{$\approx 0$} & \textbf{$\approx 0$} & \textbf{$0.028$} & $0.144$ \\
SmolLM-OpenRewriteEval-360M-Inst & \textbf{$\approx 0$} & \textbf{$\approx 0$} & \textbf{$\approx 0$} & \textbf{$\approx 0$} & \textbf{$\approx 0$} & \textbf{$\approx 0$} & $0.060$ & $0.378$ & $0.110$ & \textbf{$\approx 0$} \\
\bottomrule
\end{tabular}
}
\label{tab:wilcoxon_human_vs_models}
\end{table*}

%% file: p_values/table_human_vs_models_dolma_with_context.tex
\begin{table*}[t]
\centering
\caption{Wilcoxon rank-sum test $p$-values: each model vs.\ the human baseline (two-sided) on the Dolma benchmark, prompted generation with preceding context window. Columns denote chunk size (tokens).}
{\fontsize{8}{8.4}\selectfont\setlength{\tabcolsep}{4pt}
\begin{tabular}{lrrrrrrrrrr}
\toprule
\textbf{Model} & \textbf{50} & \textbf{100} & \textbf{150} & \textbf{200} & \textbf{250} & \textbf{300} & \textbf{350} & \textbf{400} & \textbf{450} & \textbf{500} \\
\midrule
SmolLM2-1.7B-Base- W/Context & \textbf{$\approx 0$} & \textbf{$\approx 0$} & \textbf{$\approx 0$} & \textbf{$\approx 0$} & \textbf{$\approx 0$} & \textbf{$\approx 0$} & \textbf{$\approx 0$} & \textbf{$\approx 0$} & \textbf{$\approx 0$} & \textbf{$\approx 0$} \\
SmolLM2-1.7B-Inst- W/Context & \textbf{$\approx 0$} & \textbf{$\approx 0$} & \textbf{$\approx 0$} & \textbf{$\approx 0$} & \textbf{$\approx 0$} & \textbf{$\approx 0$} & \textbf{$\approx 0$} & \textbf{$\approx 0$} & \textbf{$\approx 0$} & \textbf{$\approx 0$} \\
SmolLM2-360M-Base- W/Context & \textbf{$\approx 0$} & \textbf{$\approx 0$} & \textbf{$\approx 0$} & \textbf{$\approx 0$} & \textbf{$\approx 0$} & \textbf{$\approx 0$} & \textbf{$\approx 0$} & \textbf{$\approx 0$} & \textbf{$\approx 0$} & \textbf{$\approx 0$} \\
SmolLM2-360M-Inst- W/Context & \textbf{$\approx 0$} & \textbf{$\approx 0$} & \textbf{$\approx 0$} & \textbf{$\approx 0$} & \textbf{$\approx 0$} & \textbf{$\approx 0$} & \textbf{$\approx 0$} & \textbf{$\approx 0$} & \textbf{$\approx 0$} & \textbf{$\approx 0$} \\
\midrule
SmolLM-1.7B-Base- W/Context & \textbf{$\approx 0$} & \textbf{$\approx 0$} & \textbf{$\approx 0$} & \textbf{$\approx 0$} & \textbf{$\approx 0$} & \textbf{$\approx 0$} & \textbf{$\approx 0$} & \textbf{$\approx 0$} & \textbf{$\approx 0$} & $0.084$ \\
SmolLM-1.7B-Inst- W/Context & \textbf{$\approx 0$} & \textbf{$\approx 0$} & \textbf{$\approx 0$} & \textbf{$\approx 0$} & \textbf{$\approx 0$} & \textbf{$0.009$} & \textbf{$0.021$} & \textbf{$0.021$} & $0.052$ & \textbf{$0.034$} \\
SmolLM-360M-Base- W/Context & \textbf{$\approx 0$} & \textbf{$\approx 0$} & \textbf{$\approx 0$} & \textbf{$\approx 0$} & \textbf{$\approx 0$} & \textbf{$\approx 0$} & \textbf{$\approx 0$} & \textbf{$\approx 0$} & \textbf{$\approx 0$} & $0.090$ \\
SmolLM-360M-Inst- W/Context & \textbf{$\approx 0$} & \textbf{$\approx 0$} & \textbf{$0.031$} & $0.155$ & $0.117$ & $0.389$ & $0.533$ & $0.718$ & $0.216$ & \textbf{$0.006$} \\
\bottomrule
\end{tabular}
}
\label{tab:wilcoxon_human_vs_models_dolma_ctx}
\end{table*}

%% file: p_values/table_human_vs_models_dolma_no_context.tex
\begin{table*}[t]
\centering
\caption{Wilcoxon rank-sum test $p$-values: each model vs.\ the human baseline (two-sided) on the Dolma benchmark, unprompted generation without context window. Columns denote chunk size (tokens).}
{\small\setlength{\tabcolsep}{4pt}
\begin{tabular}{lrrrrrrrrrr}
\toprule
\textbf{Model} & \textbf{50} & \textbf{100} & \textbf{150} & \textbf{200} & \textbf{250} & \textbf{300} & \textbf{350} & \textbf{400} & \textbf{450} & \textbf{500} \\
\midrule
SmolLM2-1.7B-Base & \textbf{$\approx 0$} & \textbf{$\approx 0$} & \textbf{$\approx 0$} & \textbf{$\approx 0$} & \textbf{$\approx 0$} & \textbf{$\approx 0$} & \textbf{$\approx 0$} & \textbf{$\approx 0$} & \textbf{$\approx 0$} & \textbf{$\approx 0$} \\
SmolLM2-1.7B-Inst & \textbf{$\approx 0$} & \textbf{$\approx 0$} & \textbf{$\approx 0$} & \textbf{$\approx 0$} & \textbf{$\approx 0$} & \textbf{$\approx 0$} & \textbf{$\approx 0$} & \textbf{$\approx 0$} & \textbf{$\approx 0$} & \textbf{$\approx 0$} \\
SmolLM2-360M-Base & \textbf{$\approx 0$} & \textbf{$\approx 0$} & \textbf{$\approx 0$} & \textbf{$\approx 0$} & \textbf{$\approx 0$} & \textbf{$\approx 0$} & \textbf{$\approx 0$} & \textbf{$\approx 0$} & \textbf{$\approx 0$} & \textbf{$\approx 0$} \\
SmolLM2-360M-Inst & \textbf{$\approx 0$} & \textbf{$\approx 0$} & \textbf{$\approx 0$} & \textbf{$\approx 0$} & \textbf{$\approx 0$} & \textbf{$\approx 0$} & \textbf{$\approx 0$} & \textbf{$\approx 0$} & \textbf{$\approx 0$} & \textbf{$\approx 0$} \\
\midrule
SmolLM-1.7B-Base & \textbf{$\approx 0$} & \textbf{$\approx 0$} & \textbf{$\approx 0$} & \textbf{$\approx 0$} & \textbf{$\approx 0$} & \textbf{$\approx 0$} & \textbf{$\approx 0$} & \textbf{$\approx 0$} & \textbf{$\approx 0$} & \textbf{$\approx 0$} \\
SmolLM-1.7B-Inst & \textbf{$\approx 0$} & \textbf{$\approx 0$} & \textbf{$\approx 0$} & \textbf{$\approx 0$} & \textbf{$\approx 0$} & \textbf{$\approx 0$} & \textbf{$\approx 0$} & \textbf{$\approx 0$} & \textbf{$\approx 0$} & \textbf{$\approx 0$} \\
SmolLM-360M-Base & \textbf{$\approx 0$} & \textbf{$\approx 0$} & \textbf{$\approx 0$} & \textbf{$\approx 0$} & \textbf{$\approx 0$} & \textbf{$\approx 0$} & \textbf{$\approx 0$} & \textbf{$\approx 0$} & \textbf{$\approx 0$} & \textbf{$\approx 0$} \\
SmolLM-360M-Inst & \textbf{$\approx 0$} & \textbf{$\approx 0$} & \textbf{$\approx 0$} & \textbf{$\approx 0$} & \textbf{$\approx 0$} & \textbf{$\approx 0$} & \textbf{$\approx 0$} & \textbf{$\approx 0$} & \textbf{$\approx 0$} & \textbf{$\approx 0$} \\
\bottomrule
\end{tabular}
}
\label{tab:wilcoxon_human_vs_models_dolma_no_ctx}
\end{table*}

%% file: p_values/table_360M_vs_1p7B.tex
\begin{table*}[t]
\centering
\caption{Wilcoxon rank-sum test $p$-values for the hypothesis \emph{360M calibrated similarity $<$ 1.7B calibrated similarity} (one-sided, alternative=\texttt{less}). Columns denote chunk size (tokens).}
{\fontsize{8}{8.4}\selectfont\setlength{\tabcolsep}{4pt}
\begin{tabular}{lrrrrrrrrrr}
\toprule
\textbf{Model} & \textbf{50} & \textbf{100} & \textbf{150} & \textbf{200} & \textbf{250} & \textbf{300} & \textbf{350} & \textbf{400} & \textbf{450} & \textbf{500} \\
\midrule
SmolLM2-GSM8K-Base & \textbf{$\approx 0$} & \textbf{$\approx 0$} & \textbf{$\approx 0$} & \textbf{$\approx 0$} & \textbf{$\approx 0$} & \textbf{$\approx 0$} & --- & --- & --- & --- \\
SmolLM2-GSM8K-Instruct & \textbf{$\approx 0$} & \textbf{$\approx 0$} & \textbf{$\approx 0$} & \textbf{$\approx 0$} & \textbf{$\approx 0$} & \textbf{$\approx 0$} & --- & --- & --- & --- \\
SmolLM2-TruthfulQA-Base & \textbf{$\approx 0$} & \textbf{$\approx 0$} & \textbf{$\approx 0$} & \textbf{$\approx 0$} & \textbf{$\approx 0$} & \textbf{$\approx 0$} & --- & --- & --- & --- \\
SmolLM2-TruthfulQA-Instruct & \textbf{$\approx 0$} & \textbf{$\approx 0$} & \textbf{$\approx 0$} & \textbf{$\approx 0$} & \textbf{$\approx 0$} & \textbf{$\approx 0$} & --- & --- & --- & --- \\
SmolLM2-OpenRewriteEval-Base & \textbf{$\approx 0$} & \textbf{$\approx 0$} & \textbf{$\approx 0$} & \textbf{$\approx 0$} & \textbf{$\approx 0$} & \textbf{$\approx 0$} & \textbf{$\approx 0$} & \textbf{$\approx 0$} & \textbf{$\approx 0$} & \textbf{$\approx 0$} \\
SmolLM2-OpenRewriteEval-Instruct & $0.572$ & $0.551$ & $0.418$ & $0.567$ & $0.565$ & $0.492$ & $0.544$ & $0.494$ & $0.467$ & $0.359$ \\
\midrule
SmolLM-GSM8K-Base & $0.979$ & $0.746$ & $0.282$ & $0.266$ & $0.228$ & $0.231$ & --- & --- & --- & --- \\
SmolLM-GSM8K-Instruct & $0.145$ & \textbf{$0.014$} & $0.058$ & $0.156$ & $0.123$ & $0.373$ & --- & --- & --- & --- \\
SmolLM-TruthfulQA-Base & \textbf{$\approx 0$} & \textbf{$0.006$} & \textbf{$0.011$} & \textbf{$0.010$} & \textbf{$0.017$} & \textbf{$0.011$} & --- & --- & --- & --- \\
SmolLM-TruthfulQA-Instruct & $0.956$ & $0.928$ & $0.984$ & $0.988$ & $0.942$ & $0.919$ & --- & --- & --- & --- \\
SmolLM-OpenRewriteEval-Base & $1.000$ & $0.997$ & $0.961$ & $0.970$ & $0.943$ & $0.881$ & $0.830$ & $0.816$ & $0.854$ & $0.736$ \\
SmolLM-OpenRewriteEval-Instruct & \textbf{$\approx 0$} & \textbf{$\approx 0$} & \textbf{$\approx 0$} & \textbf{$\approx 0$} & \textbf{$\approx 0$} & \textbf{$\approx 0$} & \textbf{$0.003$} & \textbf{$0.005$} & \textbf{$0.003$} & \textbf{$\approx 0$} \\
\bottomrule
\end{tabular}
}

\label{tab:wilcoxon_360M_vs_1p7B}
\end{table*}

%% file: p_values/table_360M_vs_1p7B_dolma.tex
\begin{table*}[t]
\centering
\caption{Wilcoxon rank-sum test $p$-values for \emph{360M calibrated similarity $<$ 1.7B calibrated similarity} (one-sided) on the Dolma benchmark (unprompted generation, no context). Columns denote chunk size (tokens).}
{\small\setlength{\tabcolsep}{4pt}
\begin{tabular}{lrrrrrrrrrr}
\toprule
\textbf{Model} & \textbf{50} & \textbf{100} & \textbf{150} & \textbf{200} & \textbf{250} & \textbf{300} & \textbf{350} & \textbf{400} & \textbf{450} & \textbf{500} \\
\midrule
SmolLM2-Base & $1.000$ & $1.000$ & $1.000$ & $1.000$ & $1.000$ & $1.000$ & $1.000$ & $1.000$ & $1.000$ & $1.000$ \\
SmolLM2-Instruct & $1.000$ & $1.000$ & $1.000$ & $1.000$ & $1.000$ & $0.992$ & $0.597$ & \textbf{$0.044$} & \textbf{$0.033$} & $1.000$ \\
\midrule
SmolLM-Base & \textbf{$\approx 0$} & \textbf{$\approx 0$} & \textbf{$0.004$} & \textbf{$0.008$} & $0.133$ & $0.291$ & $0.370$ & $0.525$ & $0.105$ & \textbf{$\approx 0$} \\
SmolLM-Instruct & \textbf{$\approx 0$} & \textbf{$\approx 0$} & \textbf{$\approx 0$} & \textbf{$\approx 0$} & \textbf{$0.004$} & \textbf{$0.016$} & \textbf{$0.001$} & \textbf{$\approx 0$} & \textbf{$\approx 0$} & \textbf{$0.022$} \\
\bottomrule
\end{tabular}
}

\label{tab:wilcoxon_360M_vs_1p7B_dolma}
\end{table*}

%% file: p_values/table_instruct_vs_base.tex
\begin{table*}[t]
\centering
\caption{Wilcoxon rank-sum test $p$-values for the hypothesis \emph{Instruct calibrated similarity $<$ Base calibrated similarity} (one-sided, alternative=\texttt{less}). Columns denote chunk size (tokens).}
{\fontsize{8}{8.4}\selectfont\setlength{\tabcolsep}{4pt}
\begin{tabular}{lrrrrrrrrrr}
\toprule
\textbf{Model} & \textbf{50} & \textbf{100} & \textbf{150} & \textbf{200} & \textbf{250} & \textbf{300} & \textbf{350} & \textbf{400} & \textbf{450} & \textbf{500} \\
\midrule
SmolLM2-GSM8K-1.7B & \textbf{$\approx 0$} & \textbf{$\approx 0$} & \textbf{$\approx 0$} & \textbf{$\approx 0$} & \textbf{$\approx 0$} & \textbf{$\approx 0$} & --- & --- & --- & --- \\
SmolLM2-GSM8K-360M & \textbf{$\approx 0$} & \textbf{$\approx 0$} & \textbf{$\approx 0$} & \textbf{$\approx 0$} & \textbf{$\approx 0$} & \textbf{$\approx 0$} & --- & --- & --- & --- \\
SmolLM2-TruthfulQA-1.7B & \textbf{$\approx 0$} & \textbf{$\approx 0$} & \textbf{$\approx 0$} & \textbf{$\approx 0$} & \textbf{$\approx 0$} & \textbf{$\approx 0$} & --- & --- & --- & --- \\
SmolLM2-TruthfulQA-360M & \textbf{$\approx 0$} & \textbf{$\approx 0$} & \textbf{$\approx 0$} & \textbf{$\approx 0$} & \textbf{$\approx 0$} & \textbf{$\approx 0$} & --- & --- & --- & --- \\
SmolLM2-OpenRewriteEval-1.7B & \textbf{$\approx 0$} & \textbf{$\approx 0$} & \textbf{$\approx 0$} & \textbf{$\approx 0$} & \textbf{$\approx 0$} & \textbf{$\approx 0$} & \textbf{$\approx 0$} & \textbf{$\approx 0$} & \textbf{$\approx 0$} & \textbf{$\approx 0$} \\
SmolLM2-OpenRewriteEval-360M & \textbf{$\approx 0$} & \textbf{$\approx 0$} & \textbf{$\approx 0$} & \textbf{$\approx 0$} & \textbf{$\approx 0$} & \textbf{$\approx 0$} & \textbf{$\approx 0$} & \textbf{$\approx 0$} & \textbf{$\approx 0$} & \textbf{$\approx 0$} \\
\midrule
SmolLM-GSM8K-1.7B & $1.000$ & $0.982$ & $0.862$ & $0.719$ & $0.826$ & $0.681$ & --- & --- & --- & --- \\
SmolLM-GSM8K-360M & $0.298$ & $0.066$ & $0.335$ & $0.412$ & $0.449$ & $0.680$ & --- & --- & --- & --- \\
SmolLM-TruthfulQA-1.7B & \textbf{$\approx 0$} & \textbf{$\approx 0$} & \textbf{$\approx 0$} & \textbf{$\approx 0$} & \textbf{$\approx 0$} & \textbf{$\approx 0$} & --- & --- & --- & --- \\
SmolLM-TruthfulQA-360M & \textbf{$\approx 0$} & \textbf{$\approx 0$} & \textbf{$\approx 0$} & \textbf{$\approx 0$} & \textbf{$\approx 0$} & \textbf{$\approx 0$} & --- & --- & --- & --- \\
SmolLM-OpenRewriteEval-1.7B & $1.000$ & $1.000$ & $1.000$ & $1.000$ & $1.000$ & $0.883$ & $0.597$ & $0.179$ & $0.879$ & $1.000$ \\
SmolLM-OpenRewriteEval-360M & $0.958$ & $0.968$ & $0.872$ & $0.310$ & $0.089$ & \textbf{$0.003$} & \textbf{$0.002$} & \textbf{$\approx 0$} & \textbf{$0.009$} & $1.000$ \\
\bottomrule
\end{tabular}
}

\label{tab:wilcoxon_instruct_vs_base}
\end{table*}

%% file: p_values/table_instruct_vs_base_dolma.tex
\begin{table*}[t]
\centering
\caption{Wilcoxon rank-sum test $p$-values for \emph{Instruct calibrated similarity $<$ Base calibrated similarity} (one-sided) on the Dolma benchmark (unprompted generation, no context). Columns denote chunk size (tokens).}
{\small\setlength{\tabcolsep}{4pt}
\begin{tabular}{lrrrrrrrrrr}
\toprule
\textbf{Model} & \textbf{50} & \textbf{100} & \textbf{150} & \textbf{200} & \textbf{250} & \textbf{300} & \textbf{350} & \textbf{400} & \textbf{450} & \textbf{500} \\
\midrule
SmolLM2-1.7B & \textbf{$\approx 0$} & $0.053$ & $0.826$ & $0.424$ & $0.256$ & $0.261$ & $0.061$ & \textbf{$0.017$} & $0.977$ & $1.000$ \\
SmolLM2-360M & \textbf{$0.002$} & $0.051$ & \textbf{$0.004$} & \textbf{$\approx 0$} & \textbf{$\approx 0$} & \textbf{$\approx 0$} & \textbf{$\approx 0$} & \textbf{$\approx 0$} & \textbf{$\approx 0$} & $1.000$ \\
\midrule
SmolLM-1.7B & $0.133$ & \textbf{$\approx 0$} & \textbf{$\approx 0$} & \textbf{$\approx 0$} & \textbf{$\approx 0$} & \textbf{$\approx 0$} & \textbf{$\approx 0$} & \textbf{$\approx 0$} & $0.911$ & $0.755$ \\
SmolLM-360M & \textbf{$0.002$} & \textbf{$\approx 0$} & \textbf{$\approx 0$} & \textbf{$\approx 0$} & \textbf{$\approx 0$} & \textbf{$\approx 0$} & \textbf{$\approx 0$} & \textbf{$\approx 0$} & $0.086$ & $0.995$ \\
\bottomrule
\end{tabular}
}
\label{tab:wilcoxon_instruct_vs_base_dolma}
\end{table*}

%% file: p_values/table_cross_granularity_dolma.tex
\begin{table*}[t]
\centering

\caption{Wilcoxon rank-sum test $p$-values for the hypothesis \emph{longer-chunk calibrated similarity $<$ shorter-chunk calibrated similarity} (one-sided) on Dolma. A significant result supports Observation~3 (novelty increases with generation length).}
{\fontsize{8}{8.4}\selectfont\setlength{\tabcolsep}{4pt}
\begin{minipage}{0.48\textwidth}
\centering
\textbf{SmolLM2 (no context)}\\[4pt]
\begin{tabular}{lrr}
\toprule
\textbf{Model} & \textbf{450 vs.\ 250} & \textbf{250 vs.\ 50} \\
\midrule
1.7B-Base & \textbf{$\approx 0$} & \textbf{$\approx 0$} \\
1.7B-Inst & \textbf{$\approx 0$} & \textbf{$\approx 0$} \\
360M-Base & \textbf{$\approx 0$} & \textbf{$\approx 0$} \\
360M-Inst & \textbf{$\approx 0$} & \textbf{$\approx 0$} \\
human-Base & $0.213$ & $0.475$ \\
\bottomrule
\end{tabular}
\end{minipage}
\hfill
\begin{minipage}{0.48\textwidth}
\centering
\textbf{SmolLM2 (with context)}\\[4pt]
\begin{tabular}{lrr}
\toprule
\textbf{Model} & \textbf{450 vs.\ 250} & \textbf{250 vs.\ 50} \\
\midrule
1.7B-Base- W/Context & \textbf{$\approx 0$} & $0.078$ \\
1.7B-Inst- W/Context & \textbf{$\approx 0$} & $0.090$ \\
360M-Base- W/Context & \textbf{$\approx 0$} & $0.147$ \\
360M-Inst- W/Context & \textbf{$\approx 0$} & $0.258$ \\
human-Base- W/Context & $0.196$ & $0.454$ \\
\bottomrule
\end{tabular}
\end{minipage}
\\[12pt]
\begin{minipage}{0.48\textwidth}
\centering
\textbf{SmolLM (no context)}\\[4pt]
\begin{tabular}{lrr}
\toprule
\textbf{Model} & \textbf{450 vs.\ 250} & \textbf{250 vs.\ 50} \\
\midrule
1.7B-Base & \textbf{$\approx 0$} & \textbf{$\approx 0$} \\
1.7B-Inst & \textbf{$\approx 0$} & \textbf{$\approx 0$} \\
360M-Base & \textbf{$\approx 0$} & \textbf{$\approx 0$} \\
360M-Inst & \textbf{$\approx 0$} & \textbf{$\approx 0$} \\
human-Base & $0.213$ & $0.475$ \\
\bottomrule
\end{tabular}
\end{minipage}
\hfill
\begin{minipage}{0.48\textwidth}
\centering
\textbf{SmolLM (with context)}\\[4pt]
\begin{tabular}{lrr}
\toprule
\textbf{Model} & \textbf{450 vs.\ 250} & \textbf{250 vs.\ 50} \\
\midrule
1.7B-Base- W/Context & \textbf{$\approx 0$} & $0.187$ \\
1.7B-Inst- W/Context & \textbf{$\approx 0$} & \textbf{$\approx 0$} \\
360M-Base- W/Context & \textbf{$\approx 0$} & \textbf{$0.036$} \\
360M-Inst- W/Context & \textbf{$\approx 0$} & \textbf{$\approx 0$} \\
human-Base- W/Context & $0.196$ & $0.454$ \\
\bottomrule
\end{tabular}
\end{minipage}
}
\\[12pt]
\label{tab:wilcoxon_cross_granularity_dolma}
\end{table*}

%% file: p_values/table_prompted_vs_unprompted_dolma_smollm.tex
\begin{table*}[t]

\centering

\caption{Wilcoxon rank-sum test $p$-values for the hypothesis \emph{prompted calibrated similarity $<$ unprompted calibrated similarity} (one-sided, alternative=\texttt{less}) on the Dolma benchmark (SmolLM). Columns denote chunk size (tokens).}
{\small\setlength{\tabcolsep}{4pt}

\begin{tabular}{lrrrrrrrrrr}

\toprule

\textbf{Model} & \textbf{50} & \textbf{100} & \textbf{150} & \textbf{200} & \textbf{250} & \textbf{300} & \textbf{350} & \textbf{400} & \textbf{450} & \textbf{500} \\

\midrule

1.7B-Base & \textbf{$\approx 0$} & \textbf{$\approx 0$} & \textbf{$\approx 0$} & \textbf{$\approx 0$} & \textbf{$\approx 0$} & \textbf{$\approx 0$} & \textbf{$\approx 0$} & \textbf{$\approx 0$} & \textbf{$\approx 0$} & \textbf{$\approx 0$} \\

1.7B-Inst & \textbf{$\approx 0$} & \textbf{$\approx 0$} & \textbf{$\approx 0$} & \textbf{$\approx 0$} & \textbf{$\approx 0$} & \textbf{$\approx 0$} & \textbf{$\approx 0$} & \textbf{$\approx 0$} & \textbf{$\approx 0$} & \textbf{$\approx 0$} \\

360M-Base & \textbf{$\approx 0$} & \textbf{$\approx 0$} & \textbf{$\approx 0$} & \textbf{$\approx 0$} & \textbf{$\approx 0$} & \textbf{$\approx 0$} & \textbf{$\approx 0$} & \textbf{$\approx 0$} & \textbf{$\approx 0$} & \textbf{$\approx 0$} \\

360M-Inst & \textbf{$\approx 0$} & \textbf{$\approx 0$} & \textbf{$\approx 0$} & \textbf{$\approx 0$} & \textbf{$\approx 0$} & \textbf{$\approx 0$} & \textbf{$\approx 0$} & \textbf{$\approx 0$} & \textbf{$\approx 0$} & \textbf{$\approx 0$} \\

human-Base & $0.580$ & $0.159$ & $0.302$ & $0.202$ & $0.527$ & $0.428$ & $0.605$ & $0.399$ & $0.532$ & $0.516$ \\

\bottomrule

\end{tabular}

}
\label{tab:wilcoxon_prompted_vs_unprompted_dolma_smollm}
\end{table*}

%% file: p_values/table_prompted_vs_unprompted_dolma_smollm2.tex
\begin{table*}[t]
\centering
\caption{Wilcoxon rank-sum test $p$-values for the hypothesis \emph{prompted calibrated similarity $<$ unprompted calibrated similarity} (one-sided, alternative=\texttt{less}) on the Dolma benchmark (SmolLM2). Columns denote chunk size (tokens).}
{\small\setlength{\tabcolsep}{4pt}
\begin{tabular}{lrrrrrrrrrr}
\toprule
\textbf{Model} & \textbf{50} & \textbf{100} & \textbf{150} & \textbf{200} & \textbf{250} & \textbf{300} & \textbf{350} & \textbf{400} & \textbf{450} & \textbf{500} \\
\midrule
1.7B-Base & \textbf{$\approx 0$} & \textbf{$\approx 0$} & \textbf{$\approx 0$} & \textbf{$\approx 0$} & \textbf{$\approx 0$} & \textbf{$\approx 0$} & \textbf{$\approx 0$} & \textbf{$\approx 0$} & \textbf{$\approx 0$} & \textbf{$\approx 0$} \\
1.7B-Inst & \textbf{$\approx 0$} & \textbf{$\approx 0$} & \textbf{$\approx 0$} & \textbf{$\approx 0$} & \textbf{$\approx 0$} & \textbf{$\approx 0$} & \textbf{$\approx 0$} & \textbf{$\approx 0$} & \textbf{$\approx 0$} & \textbf{$\approx 0$} \\
360M-Base & \textbf{$\approx 0$} & \textbf{$\approx 0$} & \textbf{$\approx 0$} & \textbf{$\approx 0$} & \textbf{$\approx 0$} & \textbf{$\approx 0$} & \textbf{$\approx 0$} & \textbf{$\approx 0$} & \textbf{$\approx 0$} & \textbf{$\approx 0$} \\
360M-Inst & \textbf{$\approx 0$} & \textbf{$\approx 0$} & \textbf{$\approx 0$} & \textbf{$\approx 0$} & \textbf{$\approx 0$} & \textbf{$\approx 0$} & \textbf{$\approx 0$} & $0.517$ & \textbf{$0.029$} & \textbf{$\approx 0$} \\
human-Base & $0.580$ & $0.159$ & $0.302$ & $0.202$ & $0.527$ & $0.428$ & $0.605$ & $0.399$ & $0.532$ & $0.516$ \\
\bottomrule
\end{tabular}
}
\label{tab:wilcoxon_prompted_vs_unprompted_dolma_smollm2}
\end{table*}